\documentclass{article}

     \PassOptionsToPackage{numbers, compress}{natbib}
     \usepackage[preprint]{neurips_2025}

\usepackage[utf8]{inputenc} % allow utf-8 input
\usepackage[T1]{fontenc}    % use 8-bit T1 fonts
\usepackage{hyperref}       % hyperlinks
\usepackage{url}            % simple URL typesetting
\usepackage{booktabs}       % professional-quality tables
\usepackage{amsfonts}       % blackboard math symbols
\usepackage{nicefrac}       % compact symbols for 1/2, etc.
\usepackage{microtype}      % microtypography
\usepackage{xcolor}         % colors
\usepackage{microtype}
\usepackage{graphicx}
\usepackage{array}
\usepackage{subfigure}
\usepackage{amsmath}   %  \tfrac, \dfrac, \operatorname, \tag, align, …
\usepackage{amssymb}   %  \mathbb, \mathcal, \leqslant, …
\usepackage{bm}        %  \bm   (bold math symbols)
\usepackage{algorithm}
\usepackage{algorithmic}
\usepackage{makecell}
\usepackage{multirow}

\title{SifterNet: A Generalized and Model-Agnostic Trigger Purification Approach}

\author{%
 Shaoye Luo$^{1,2}$ \quad Xinxin Fan$^{1,2}$\thanks{ Corresponding author} \quad Quanliang Jing$^{1}$ \quad Chi Lin$^3$ \quad Mengfan Li$^{1,2}$ \quad \\ \textbf{Yunfeng Lu$^{4*}$ \quad Yongjun Xu$^{1,2}$} \\
$^1$Institute of Computing Technology, Chinese Academy of Sciences \\
\quad $^2$University of the Chinese Academy of Sciences \\
\quad $^3$Dalian University of Technology 
\quad $^4$Beihang University \\
\texttt{\{fanxinxin,limengfan22z,luoshaoye24s,xyj\}@ict.ac.cn}\\
\texttt{c.lin@dlut.edu.cn} \quad 
\texttt{lyf@buaa.edu.cn}
}

\begin{document}

	\maketitle

\begin{abstract}
Aiming at resisting backdoor attacks in convolution neural networks and vision Transformer-based large model, this paper proposes a generalized and model-agnostic trigger-purification approach resorting to the classic Ising model. To date, existing trigger detection/removal studies usually require to know the detailed knowledge of target model in advance, access to a large number of clean samples or even model-retraining authorization, which brings the huge inconvenience for practical applications, especially inaccessible to target model. An ideal countermeasure ought to eliminate the implanted trigger without regarding whatever the target models are. To this end, a lightweight and black-box defense approach SifterNet is proposed through leveraging the memorization-association functionality of Hopfield network, by which the triggers of input samples can be effectively purified in a proper manner. The main novelty of our proposed approach lies in the introduction of ideology of Ising model. Extensive experiments also validate the effectiveness of our approach in terms of proper trigger purification and high accuracy achievement, and compared to the state-of-the-art baselines under several commonly-used datasets, our SiferNet has a significant superior performance.
\end{abstract}

\section{Introduction}\label{submission}
The surge of artificial intelligence is attracting more and more attention in diversities of applications, nevertheless, the security and privacy issues also accompany simultaneously, such as adversarial attacks, data poisoning, backdoor (Trojan) injection, model parameters and network structure leakage, etc. This paper focuses on the defense of backdoor attacks, and compared to adversarial attack, backdoor attack becomes more difficult to defend, deriving from its covert property, i.e., a sample with pre-implanted trigger can deceive target model to make incorrect inference, while a sample without such a trigger cannot trigger the attack, i.e. the target model would make correct inference. In recent years, as the emergence and pervasiveness of federated learning architecture, the users' datasets are decentralized locally, only with the operation to upload the well-trained parameters to the aggregator, thereby, the attacker can easily train a backdoor-integrated model through resorting to trigger-implanted training datasets, leading to the trained model getting the backdoor involved.

To date, there mainly exist two lines of work for the defense against backdoor attacks, one is from the model viewpoint, i.e. this line of work aims to eliminate triggers of poisoned samples. The recent works \cite{Neural_Cleanse, BayBFed, Demon_in_the_Variant, FLAME, Joseph23, Huang22} proposed a set of white-box-oriented countermeasures to eliminate backdoor by fine-training or trigger-rebuilding operations. These methods not only ascertain if the models are contaminated, but also remedy the neural networks to make trigger unworkable. Given the above methods are white-box based, i.e. the model's information are pre-known and accessible, thus, this is the reason why they are capable to achieve valid defense. However, such white-box-based defense methods unfortunately have limited adaption, namely they need to know details of target model in advance, on the other hand, must make specific hypothesis on attack behavior, thereby the performance would be deteriorated dramatically while encountering the variances of network structures, datasets and attack manners. 
Oppositely, some methods are black-box based \cite{Yinpeng21, DeepInspect, HaoFu23, SCALE_UP}, i.e. the detail of target model is unknown, but the defender can access it and obtain the output information as a referral for the next-step defense. Resorting to such query-based diagnosis, the model can be detected whether a backdoor is implanted into. A representative category of methods is to leverage Meta-Learning to judge whether backdoor is injected or not, that is, at first need to assume the output distributions by backdoor-injected model and benign model are different. Then, using massive output vectors inferred by both benign and backdoor-implanted models to retrain a new Meta-Classifier to infer whether the target model is clean or contaminated. Consider the requirement of a large number of retraining samples, they usually take a huge time overhead for such category of black-box-based defense mechanisms.

The other line of backdoor-defense work is from the perspective of sample data, i.e. inspect whether a trigger is injected into the clean samples, and even remedy it \cite{Februus}. Compared to the model-oriented defense, this sample data-oriented study is lightweight, namely, the existing work judges whether a sample has trigger or not through inspecting the parameters of neural network, or accomplishing the elimination of triggers. This data-oriented defense is more adaptable resulting from its capability of handling diverse models, datasets and manners of backdoor attacks. Another countermeasure is to directly divide the clean samples from contaminated ones, however, this approach usually need leverage the parameter variations to execute correct division.

Aiming to tackle the adaptability and time-consuming problems, we propose a novel generalized and model-agnostic trigger-purification approach. In a netshell, three main contributions are involved:

\begin{itemize}
	\item Inspired by the ideology of Ising model in classical physics, we propose a generalized trigger-purification scheme to eliminate the triggers of poised samples resorting to the memorization-association functionality of Hopfield network.
	\item We theoretically analyze the stability of Hopfield-network via Hebbian learning, and figure out an effective and efficient trigger-purification framework.
	\item Extensive experiments are performed to validate the effectiveness of our proposed approach, and the results showcase our approach significantly outperforms the baselines in terms of attack success rate of contaminated samples and inference accuracy.
\end{itemize}

%The rest is organized as follows: Section \ref{Sec:ProblemOverview} reviews the problems of backdoor defense and briefly state our proposed approach. Section \ref{Sec:OurMethod} firstly introduces the preliminary of Ising model and purification principle, and then presents the trigger-purification framework. Section \ref{Sec:Experiments} exhibits multi-facet experiments to verify the effectiveness of our approach. Furthermore, two skills to enhance routine accuracy are also introduced in this section. We review related work in Section \ref{Sec: RelatedWork}, discuss the limitations \ref{Sec: OpenIssue} and conclude this paper in Section \ref{Sec: Conclusion}.

\section{Problem statement and overview}\label{Sec:ProblemOverview}

\subsection{Poor generalization and high overhead}
Like white-box and black-box adversarial attacks, there also exist white-box and black-box adversarial defense. For the former, the defender requires the target model’s information \cite{Februus, SentiNet} , such as network structure, training parameters, accessibility to target model, retraining approval, etc. Moreover, the white-box defense in general need make particular assumption on the attack behaviors, such as assume the trigger used by the attacker is a symbol with clear boundary, or the trigger pattern for the model is dominant. The common operation is to fine-tune and prune the model to resist the adversarial affection of the implanted trigger. Nevertheless, this line of work not only requires huge time overhead to preform retraining, but also needs to prepare a large number of clean samples to purify the contaminated neural network. More importantly, this kind of defense mechanisms will become dramatically deteriorated once the attack manners and datasets are altered. From another angle, many white-box defense approaches usually judge backdoor relying on a mono-characteristic, thus the conversion of attack manners would decline the effect of resistance. From the discussion above, we can easily know such white-box defense is model-specific, lacking of the generalization capability towards different target models and backdoor attacks. 

For the black-box adversarial defense, although the target model’s information becomes inaccessible, however, the defender still needs to acquire permission to execute massive queries, then utilize the feedback information to train a Meta-Classifier for backdoor detection \cite{Yinpeng21, DeepInspect}, along with the assumption that the model has a different distributions from the benign counterpart. From this point, we know the generation of an accurate Meta-Classifier will depend on a large number of query-based output samples, which accordingly causes costly training overhead.

\subsection{Trigger-dependence and inaccurate location}
%Compared to the time-consuming of model-oriented defense, the data-oriented defense demands low overhead, stemming from the trigger-detecting operation only without the necessity to access target model. This line of work can also be discussed from white-box and black-box viewpoints. 

Currently, several approaches are proposed to concentrate on the breach of trigger, such as STRIP \cite{STRIP}, SentiNet \cite{SentiNet}, IFMV \cite{IFMV}, etc. The key of these trigger-based defense methods lies in: i) considering the trigger as another class/label, e.g. STRIP \cite{STRIP} recognizes the trigger as an extra class, and utilizes a small clean dataset and detector to judge whether a sample is implanted into trigger or not; ii) assuming trigger has recognizable boundary, then segment the trigger region, e.g., SentiNet \cite{SentiNet} recognizes the suspect region, which enables the model to output same inferential results while posting onto massive clean datasets, as the implanted trigger; and iii) eliminating the adversarial affection of trigger via purifying samples, e.g., IFMV \cite{IFMV} employs random noise to fuzz the contaminated samples, by which the trigger becomes unworkable. From the first two categories, we know the success of defense relies on condition that trigger has particular characteristic and boundary. Nevertheless, once the pattern of trigger changes, the resistance will become deteriorated. For the third category, random-noise addition can easily cause inaccurate location during the course of trigger elimination, especially for complex dataset, such as three-channel image data. 

In addition, we can classify the backdoor-detection methods into two categories by their application phase: i) those that perform detection and defense during the pre-training stage; and ii) those that carry out data purification and detection during the inference stage after the model has been deployed. For example, SCAN \cite{Demon_in_the_Variant} requires analysis of the entire training set to determine whether the input data contains backdoor trigger patterns, thus this type of backdoor detection is conducted in the pre-training stage. Because of the accessibility of full dataset during pre-training, discovering backdoor patterns allows to avoid attacks during training and obtain a clean model. Defense methods that perform purification and detection in the inference stage typically have a different focus. Since the available data at inference time is usually limited, statistics-based countermeasures are in general not feasible. For instance, STRIP\cite{STRIP} is designed based on the assumption the triggers are independent of the main semantic region, while TeCo\cite{TeCo} and IFMV\cite{IFMV} work by adding various perturbations to the input sample with the purpose of disrupting the backdoor structure while simultaneously preserving the original semantic content. Taking into account the complexity of backdoor defense and detection methods, we sketch a clear description from the perspectives of training/inference stage, computation complexity, and black/white-box as exhibited in Table \ref{tab:backdoor-defense-methods} in Appendix \ref{Sec:categorybackdoordefense}.

\subsection{Overview of our method}
A proper mitigation should be able to handle the drawbacks above, inspired by Ising model in classical physics, we propose a novel generalized and model-agnostic trigger-purifying approach SifterNet through leveraging memorization-association functionality of Hopfield network. SifterNet not only dramatically decreases the time overhead, but it also achieves proper performance to purify triggers while simultaneously maintaining an appropriate accuracy on clean samples. The basic principle is to utilize the “recall” property of Hopfield network, by which the target model enables to memorize the clean sample' pattern. Once a trigger-implanted sample is input into Hopfield network, the trigger can be purified by such “recall” effect. Our method is model-agnostic, and unnecessary for particular hypothesis on backdoor-attack behaviors, thus, it can be easily deployed in practice.

%Upon the inspection on both model-oriented and data-oriented backdoor defense, we think a proper countermeasure should be able to handle the aforementioned drawbacks. Keeping this in mind, inspired by Ising model in physics science, we propose a novel generalized and model-agnostic trigger-purifying approach SifterNet through resorting to the memorization-association functionality of Hopfield networks, SifterNet not only dramatically decreases time overhead, but it also achieves proper performance to purify trigger while simultaneously maintaining an appropriate accuracy on the purified samples. The core principle of our approach is to leverage the “recall” property of Hopfield network, by which the neural network enables to memorize the clean samples' patterns. Once a trigger-implanted sample is input into the trained Hopfield network, the trigger will be purified by such “recall” functionality. Therefore, our approach is model-agnostic, and unnecessary for particular hypothesis on attack behaviors, thus, it can be easily deployed in practice. 

\section{The proposed method}\label{Sec:OurMethod}
\subsection{Preliminaries}

\textbf{Research Motivation.} The backdoor purification is accomplished through resorting to Hopfield network, while Hopfield network’s idea comes from Ising model. For an Ising model with two-state spins, a.k.a. Ising-spin system, one can imagine if a particular spin is inconsistent with the directions of its neighbors, the system’s energy become high. In such a state, the spin tends to spontaneously flip to align with the majority of its neighbors, thus lowering the energy. This process can be described as Glauber dynamics in physics, the rule is to randomly choose a spin and probabilistically flip it towards the reduction of energy. In an Ising-spin system, a stable state implies the system reaches thermodynamic equilibrium, at which point the system’s energy is at a local minimum, and any individual spin flip would lead to the increase of energy, thus the system cease to change spontaneously. This stable arrangement in a simple ferromagnetic Ising model corresponds to an ordered state where all spins are aligned, or in a spin glass model with complex couplings corresponding to a spin arrangement with some intricate pattern.

The Hopfield network is mechanistically inspired by the Ising model, i.e. it uses binary neurons to simulate physical spins, symmetric connection weights to simulate spin coupling, an energy function to characterize the quality of the system state, and iterative updates to drive Hopfield network to a stable state at an energy minimum, achieving the so-called "memorizing" patterns in these stable states. For Hopfield network, a stable state refers to the network converges into an equilibrium point, that is to say, upon further update none of the neurons would change their values, which implies Hopfield network state is at a "local energy minimum" state. Consider from the "memorization" perspective, Hopfield network can store memories by setting the weights so that the anticipated patterns become the stable states. Once trained, the network forms an "energy landscape" wherein the memory patterns indicate the "valleys". When a new initial state is given, the network evolves pursuant to the update rules, and it gradually move toward the closest "valley" to this initial state, and finally settling in that "valley" without further change. In other words, the stable state to which Hopfield network converges is the stored memory pattern. Such a memory storage process reflects the spontaneous evolution toward an energy minimum, which is exactly what the black-box backdoor defense needs. On the other hand, consider from the "association" perspective, Hopfield network recalls the memory closest to the current state. In addition, as for the transformation from input samples into the "energy valleys", we here utilize the Hebb learning, which is inspired by the biological neurons.

%Our work aims to eliminating the trigger pixels, thus, we can naturally train a Hopfield network to memorize the clean-image's state, then even if it suffers backdoor to make some pixels changed, the trained Hopfield network can still restore the clean state by recalling the association function. 

\textbf{Two-Dimensional Ising Model.} %Our approach is inspired by the ideology of classic Ising model, and at first we would like to give a brief introduction. In order to investigate the complex phase transitions in ferromagnetic materials in classical physics, 
One-dimensional Ising model was originated by Wilhelm Lenz and studied by Ernst Ising \cite{Ising1925}. However, this one-dimensional Ising model cannot exhibit the phase-transition phenomena. Subsequently, the statistical physicist Lars Onsager \cite{Krcmar22, McCoy1973} proposed two-dimensional Ising model and successfully observed the phase-transition phenomena. 
%Namely, from the viewpoint of two dimension, the physicists utilize Ising model to mimic the interaction of each small magnet during the heating process as an arrangement of magnet needles pointing up or down, along with the constraint that each small magnet is influenced by its neighbors. Based on this arrangement, physicists can explain the phenomena of disappearance of magnetism at a critical temperature and reappearance below such a temperature. Given the interaction between small magnets is affected by multiple factors, such as energy, environmental thermal, etc., the magnetism would change randomly, and the degree of randomness is subject to the temperature, that is, random disturbance is enhanced at high temperature, leading to disorder state of the magnets, further giving rise to the neutralization (disappearance) of the system’s magnetism, inversely, the magnet becomes stable at low temperature, and the system displays a uniform magnetic direction, leading to clear magnetism.
Fig. \ref{fig:IsingModel} exhibits the two-dimensional microscopic scenario of a lattice arrangement wherein the magnet consists of many small magnetic needles with binary arrow directions/states: up (+1) and down (-1). As aforementioned, the interaction can be characterized by the energy. According to the principle of Ising model, we have the definition that the energy is reduced by 1 unit while the magnetic needles of two adjacent squares are in the same direction; otherwise the energy increased by 1 unit. On the other hand, assume there exists extra magnetic field that enables to affect the magnetic needles, similarly, if the magnetic needle has same direction as the outside magnetic field, then the energy will be reduced, thus it can be formulated as 
\begin{equation}
E_{\{s_i\}} = -J \sum_{\langle i,j \rangle} s_i s_j - H \sum_{i=1}^{N} s_i ,
\end{equation}
where $J$ denotes an energy-coupling constant, $<i, j>$ presents all adjacent pairs of small magnetic needles. If a pair has the same state $s_i$=$s_j$, then the energy will declined by $J$. $H$ stands for the strength of external magnetic field. If it is upward, $H$ is set to be positive; otherwise negative. If the magnetic needle aligns with the external magnetic field, then energy is declined by 1 unit. If the direction is opposite to the external, then the energy raises by 1 unit.

\begin{figure} [bthp]
\centering
\subfigure[2D Ising model with lattice arrangement]{
%	\subfloat[2D Ising model with lattice arrangement]{
	\begin{minipage}[t]{0.45\columnwidth}
		\centering
		\includegraphics[width=1.1in, height=1.0in]{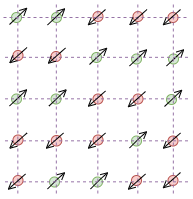}
		%	\caption{xx}
		\label{fig:IsingModel}
	\end{minipage}
}
\vspace{0.02cm}
%     \quad
\subfigure[Black-white image represented by magnetic pins]{
	%	\subfloat[Black-white image represented by magnetic pins]{
		\begin{minipage}[t]{0.5\columnwidth}
			\centering
			\includegraphics[width=1.1in, height=1.0in]{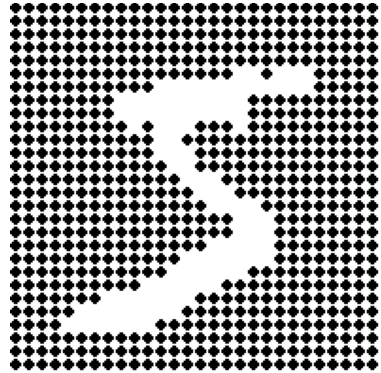}
			%	\caption{xx}
			\label{fig:IsingForImage} 
		\end{minipage}
	}
	\caption{Ising model and its represented image.}
	\label{fig:IsingModel}
\end{figure}
	
A system will tend to be stable if there is no external energy input, that is to say, the system evolves with the accompaniment of energy-reducing variation. Hence, if an external magnetic field exists, most of the small magnetic needles would align to same direction as the magnetic field. Nevertheless, in real-world scenarios, apart from the influence from the energy of the system, there also exists thermal fluctuation in noise environment, which can also affect the magnets to randomly change needles' direction. Thereby, for each small magnetic needle, it is in fact influenced by three forces: neighboring magnetic needles, external magnetic field and environment-noise disturbance. In the case of noise ignorance, each small magnetic needle tends to align to external magnetic field when the temperature close to zero, leading to the system states with all "+1" or "-1", depending on the direction of the external magnetic field parameter $H$. Another case, even with a high temperature, if the interaction strength $J$ is set to a very small value, the interaction affection between neighbors can also be ignored, which gives rise to completely random states of these small magnetic needles. Therefore, both constant $J$ and $H$ are critical parameters to decide the evolution tendency of magnetism.

\subsection{Principle of trigger-purification}
From the analysis on Ising model, we know that one key characteristic lies in the energy can naturally disperse through pairwise interaction between neighbors (magnetic needles). From the microscopic perspective, the system evolves into a stable state gradually over time. Therefore, a natural idea appears, i.e. whether a microscopic evolution regime can be established to reserve clean pattern and eliminate adversarial (trigger) pattern by optimizing the macroscopic objective (energy). Towards this end, we propose a Hopfield network-based trigger purification approach. Hopfield network \cite{Hubert21} is well-known neural network and its advantage lies in that it can memorize specific pattern through training and recalling original pattern whenever needed. In detail, it can store memory pattern through adjusting the weights of connections between neurons, and then set the pattern as the lowest-energy state. Under adequate conditions, i.e. coordinate energy-coupling constant $J$ and external magnetic field strength $H$, Hopfield network can naturally evolve into the lowest-energy state via the aforementioned interaction rule of Ising model. Intuitively, we can refer to the magnetic needles as neurons, and accordingly endow each neuron two states: activated (+1) and deactivated (-1), and denote the weight over each connection from neuron $i$ to neuron $j$ as $w_{ij}$. For the details, at first each neuron is endowed an activated state or a deactivated state, then we update its state during the training procedure with the following rules:
\begin{equation} \label{Equ-3}
s_i(t+1) =
\begin{cases} 
1 & \text{if } \sum\limits_{j} w_{ij} s_j(t) > \tau_i \\
-1 & \text{otherwise}
\end{cases} ,
\end{equation}
where $\tau_i$ is the threshold to control the state variation of neuron $i$, that is to say, for a neuron, if all the neighboring neurons are activated, it would be likely to be activated normally. From the viewpoint of small magnetic needles in Ising model, it is equivalent to minimize the global energy function with spin alignment. Given a finer-grained consideration of energy-coupling constant and magnetic field strength, we respectively replace constant $J$ with pairwise weight, and $H$ with activation threshold. Thus, the energy function can be redefined as
\begin{equation} \label{Equ-Energy}
E_{\{s_i\}} = -\sum_{\langle i,j \rangle} w_{ij} s_i s_j - \sum_{i=1}^{N} \tau_i s_i .
\end{equation}

Although the principle of Hopfield network inherits from Ising model, it indeed brings a microscopic control on neurons using two variables $w_{ij}$ and $\tau_i$ instead of the two constants, in this way, the interaction strength is dependent on each connection's weight over a pair of neurons, and the magnetic field strength also varies from neuron to neuron. Since many objects can be described by modeling two states, Hopfield network can handle more practical objects like black and white photos. As shown in Fig. \ref{fig:IsingForImage}, we can use 1 and -1 to represent the white pixel and black pixel.

\textbf{Memorization-Association Functionality.} Our work aims to make Hopfield network remember the clean pattern rather than the adversarial, i.e. achieve "memorization-associative" functionality with the purpose of cleaning adversarial pattern through training Hopfield network with appropriate connection weight $w_{ij}$ over pairwise neurons. The training ("memory") phase of Hopfield network mainly uses Hebbian learning rule, proposed by Donald Hebb to describe how the behaviors of neurons influence the pairwise connection. The core idea is that if two neurons are activated simultaneously, their correlation is relatively strong, and their connection weight should be strengthened. On the other hand, if one neuron is activated but the other connected neuron is not activated, the correlation will weaken, and the strength of their connection weight should be correspondingly reduced. If we want to memorize multiple states (vectors), we only need to add the weight matrices trained from each state to obtain the final weight matrix. 

Assume Hopfield network wants to memorize a set of vectors $
\left\{ {V_1 ,_{} V_2 ,_{}  \cdots ,_{} V_m } \right\} $, each of which can be defined as $V_i$=$<$1, -1, 1, $\cdots$, 1$>$ with $n$ dimensions, and updated according to the rule:
\begin{equation} \label{Equ-4}
w_{ij} = \sum_{v=1}^{m} V_v(i) V_v(j) ,
\end{equation}
where $V_v(i)$ indicates the value of the $i$-th component of the $V$-th vector that needs to learn. After training, a set of weights can be generated, by which any input data as the initial state of neuron can be asymptotically converged to one of the memorized vectors {$V_1$, $V_2$, $\cdots$, $V_m$} \cite{Hebb05, Hopfield1982, Hopfield1984}.  

\begin{figure}[tbhp]
\centering
\includegraphics[width=4.6in]{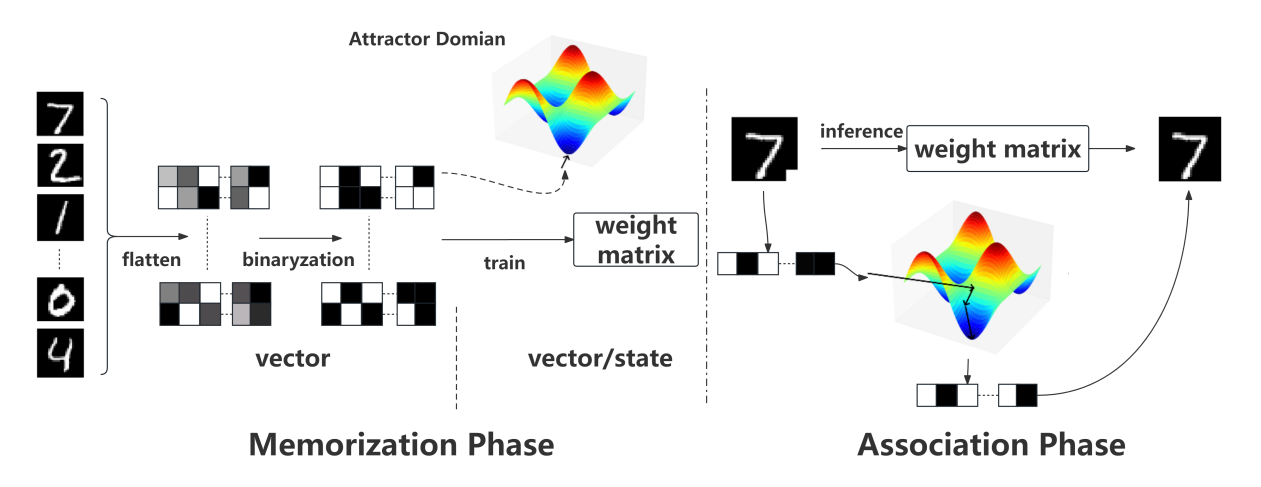}
\caption{Memorization-association functionality of Hopfield network.}
\label{fig:Memorization-Association}
\end{figure}

In order to accomplish the trigger purification, Fig. \ref{fig:Memorization-Association} exhibits the two-phase functionalities of Hopfield network: i) \textbf{Memorization.} It is also recognized as a training phase, i.e. Hopfield network modifies the weight value over each connection using Formula \ref{Equ-4}. Here we can assume there is no influence of extra magnetic field, i.e. set $\tau$ =0, and only consider the interactive influence among neurons. In this case, according to Formula \ref{Equ-Energy}, we know that if $w_{ij}$ is equal to $s_i s_j$, then, the system’s energy would be minimum at such state, which is exactly the anticipated state what we want Hopfield network to memorize; and ii) \textbf{Association.} It is an evolution procedure according to the trained weight in the memorization phase, namely, when a vector is input into the trained Hopfield network, it will be associated to the most similar (closest) memorized vector via the formula:
\begin{equation}
U_i = \sum_{j} V_j w_{ji} + \tau_i ,
\end{equation}
where the symbol $U_i$ denotes the $i$-th component of the associated vector.
In general, Hopfield network can also be recognized as an energy system wherein multiple energy-minimum states exist, i.e. so-called "attractors", and the set of all points that can reach these attractors along the direction of energy reduction is defined as "attractor domain". From the angle of trigger purification, the memorization course is to set the attractor to the position of clean samples that ought to be memorized, by which the trigger-purification can be achieved resorting to the association function. In this way, a trigger-implanted sample can be appropriately recovered if it falls into the "attractor domain".%, this phenomenon would happen due to the energy-minimum constraint by Hopfield network.

%\subsection{Stability Warrant with Hebbian Learning}
\textbf{Stability Analysis with Hebbian Learning.} In addition, to guarantee the stability requirement of Hopfield network, we also provide the formal analysis resorting to Hebbian learning in Appendix \ref{stabilityHebbian}.

\subsection{Trigger-purification framework}
\begin{figure*}[htbp]
	\vskip 0.2in
	\begin{center}
		\centering
		\includegraphics[width=4.6in, height=1.45in]{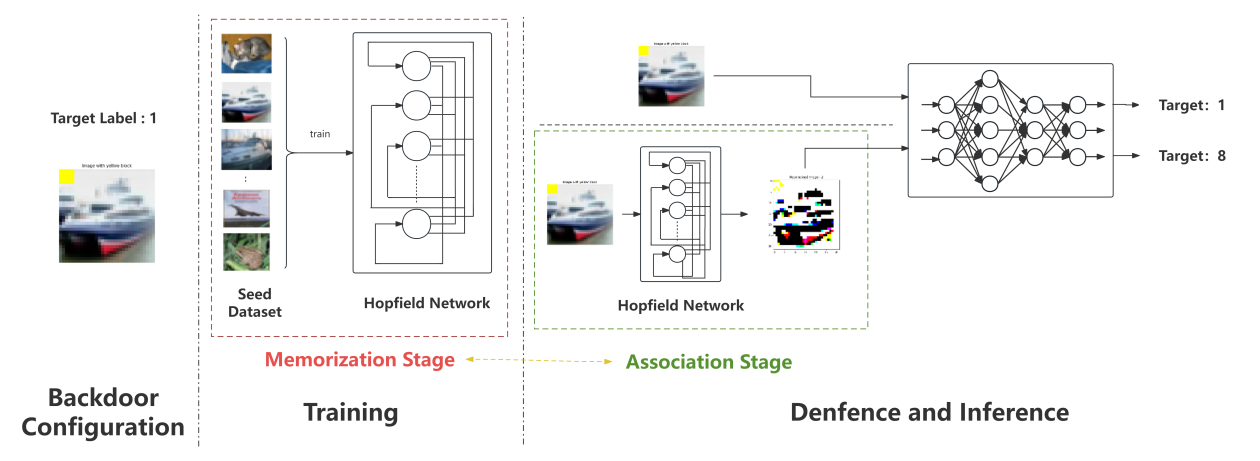}
		\caption{Overall framework of trigger-purification.}
		\label{fig:purificationFramework}
	\end{center}
	\vskip -0.2in
\end{figure*}

As shown in Fig. \ref{fig:purificationFramework}, the overall framework of trigger-purification method involves four components: i) Seed-Dataset Selection; ii) Hopfield Network Training; iii) Trigger-Implanted Sample Purification; and iv) Model Inference Towards Purified Sample. %Next, we detail each component as follows.

\textbf{Pipeline.} As first, a clean seed dataset covering all categories of samples is selected to train Hopfield network. The subsequent experimental results showcase each category does not need too many seed samples, as long as it warrants the trained Hopfield network to memorize all categories of samples in the training dataset. For a trigger-implanted sample, if its clean pattern is memorized during the training phase of Hopfield network, once it goes through the trained Hopfield network, the trigger would be purified appropriately. 

Given Hopfield network originates from Ising model, thus there only exist two states for each neuron, i.e. activated and deactivated. In other words, Hopfield network can only handle the binary-value sample, i.e. purely black and white image. There is no problem for the single-channel grayscale image like MNIST dataset \cite{LeCun1998}, the key lies in mapping pixel $p$ in range [0, 255] into interval [-1, 1] and simultaneously warranting a low precision loss. The concrete mapping formula can be set as $y$=(2*$p$)/255-1, then, referring to a predefined threshold, set it as +1 if larger than the threshold, otherwise set as -1. This mapping manner is valid for those simple-semantics samples, but unsuitable for those sophisticated-semantics samples, such as FashionMNIST \cite{Fashion_MNIST}. Thereby, to handle this plight, we employ another skill-localized differentiation \cite{Otsu1979} to binarize the grayscale image into a purely black and white image. This localized differentiation is an image-conversion technique especially for those images with uneven lighting or complex backgrounds. Its basic principle is to determine each pixel's mapping value based on its surrounding neighbors' pixels. For detail, an average filter $A$ with kernel $k$ is defined, and it is a matrix $K*K$, all of whose elements are equal to the value $1/k^2$. Assume $I$ is the original image, then the local average image $M$ can be presented as a two-dimensional convolution of matrix $I$ and matrix $A$. The convolution's boundary processing mode is same, which implies the output image $M$ has the same size as the original image $I$, and the boundary pixels are processed using the fill of zeros. The local difference threshold enables to generate the binary image $B$ through comparing each pixel in image $I$ with its corresponding local mean image $M$. This procedure can be formulated as:
\begin{equation}
B(x, y) =
\begin{cases} 
255 & \text{if } I(x, y) > M(x, y) \\
0 & \text{otherwise}
\end{cases} ,
\end{equation}
where $B(x, y)$ is the pixel value of mapped binary image at position $(x, y)$, while $I(x, y)$ and $M(x, y)$ are the pixel values of original image and locally averaged image at the same position.

\textbf{Three-Channel Image Handling.} To further make our work applicable to the popular three-channel samples, we need elaborate the overall framework of trigger purification. In detail, for the three-channel images, three channels are separately trained, giving rise to three weight matrices after training accomplished. During the course of trigger purification, each channel goes through its corresponding trained Hopfield network, after which three single-channel images are combined to generate a three-channel image, each undergoes "recall" processing by Hopfield network. Consider the complexity of multi-step implementation of our proposed SifterNet, we provide the pseudo codes to detail the critical operations in Algorithm \ref{Algorithm-1} and Algorithm \ref{Algorithm-2} in Appendix \ref{algorithm}.

\section{Experiment evaluation} \label{Sec:Experiments}
%Prior to implementing experiments, one critical problem need to be solved, i.e. the insufficiency of capacity of Hopfield network, which usually impedes the optimal performance. To tackle this point, following previous solution, we randomly select some points to converge, in this way, it can better preserve the semantics-correct regions of the image. Moreover, take into consideration of side effect of trigger purification, we also employ a data augmentation to enhance the attack resilience of target model, that is to say, reduce the confidence of clean data misclassified by target model due to deformation caused by the trigger-purification effect of Hopfield network.
\subsection{Configuration} \label{experimentConfiguration}
%I expect to use the MNIST, CIFAR10, FashionMNIST, and GTSRB datasets for evaluation, as well as comparisons using STRIP, SentiNet, and Input Fuzzing and Majority Voting methods. The network parameters are as follows:
The configurations on the statistics of four commonly-used datasets: 
MNIST \cite{LeCun1998}, FashionMNIST \cite{Fashion_MNIST}, CIFAR-10 \cite{Krizhevsky2009} and GTSRB \cite{Houben13}, four representative baselines: SentiNet \cite{SentiNet}, STRIP \cite{STRIP}, IFMV \cite{IFMV}, and Teco \cite{TeCo}, five routine backdoor attacks: BadNet \cite{BadNets}, Input-Aware \cite{InputAware}, WaNet \cite{WaNet}, Blended \cite{Blended}, SIG \cite{SIG}, two basic evaluation metrics Accuracy ($Acc.$) and attack success rate ($ASR$), and execution environment are detailed in Appendix \ref{Sec:Configuration}. Moreover, the setting of memorization iteration is depicted in Appendix \ref{memorizationIterationSetting}, and statistics on the volume of clean samples vs. purification effectiveness is demonstrated in Appendix \ref{Sec:cleanSamplevsPurification}. 

\subsection{Performance analytics}
We in fact evaluate the effectiveness of our method using two platforms: i) our self-built neural network model; and ii) the commonly-used open-source BackdoorBench \cite{BackdoorBench}. Given the limited space, the self-built platform-oriented experiments and analytics are placed in Appendix \ref{Sec:selfBuiltPlatformExperiment}.

\begin{table}[tbhp]
	\centering
	\caption{Performance on CIFAR-10 Datasets with BackdoorBench}
	\label{tab:cifar10_results}
	\begin{small}
%	\vspace{1em}
	\begin{tabular}{lcccccc}
		\toprule
		\multirow{2}{*}{\textbf{Defense}} &
		\multicolumn{2}{c}{\textbf{BadNet}} &
		\multicolumn{2}{c}{\textbf{Input-Aware}} &
		\multicolumn{2}{c}{\textbf{WaNet}} \\
		\cmidrule(lr){2-3}\cmidrule(lr){4-5}\cmidrule(lr){6-7}
		& Acc. & ASR & Acc. & ASR & Acc. & ASR \\
		\midrule
		Clean     & 75.02\% & 95.70\% & 73.97\% & 98.51\% & 77.77\% & 83.96\% \\
		SifterNet & 60.02\%$\pm$10\% & 13.67\%$\pm$7\% & 53.05\%$\pm$4\% & 48.60\%$\pm$14\% & 55.52\%$\pm$7\% &  5.00\%$\pm$2\% \\
		IFMV      & 59.58\% & 95.92\% & 53.84\% & 98.47\% & 51.47\% &  0.50\% \\
		\midrule
		& \multicolumn{2}{c}{\textbf{AUC}} & \multicolumn{2}{c}{\textbf{AUC}} & \multicolumn{2}{c}{\textbf{AUC}} \\
		\midrule  % 中间的横线
		STRIP     & \multicolumn{2}{c}{0.957}    & \multicolumn{2}{c}{0.417}   & \multicolumn{2}{c}{0.433} \\
		TeCo      & \multicolumn{2}{c}{0.955}     & \multicolumn{2}{c}{0.937}   & \multicolumn{2}{c}{0.859} \\
		\bottomrule
	\end{tabular}
	
	\vspace{1em}
	
	%—— 第二部分（后半截列） ——
	\begin{tabular}{lcccc}
		\toprule
		\multirow{2}{*}{\textbf{Defense}} &
		\multicolumn{2}{c}{\textbf{Blended}} &
		\multicolumn{2}{c}{\textbf{SIG}} \\
		\cmidrule(lr){2-3}\cmidrule(lr){4-5}
		& Acc. & ASR & Acc. & ASR \\
		\midrule
		Clean     & 74.01\% & 82.91\% & 69.02\% & 83.26\% \\
		SifterNet & 60.01\%$\pm$3\% & 38.38\%$\pm$4\% & 60.50\%$\pm$6\% & 37.22\%$\pm$3\% \\
		IFMV      & 58.23\% & 66.12\% & 58.46\% & 83.11\% \\
		\midrule
		& \multicolumn{2}{c}{\textbf{AUC}} & \multicolumn{2}{c}{\textbf{AUC}} \\
		\midrule  % 中间的横线
		STRIP     & \multicolumn{2}{c}{0.529} 	& \multicolumn{2}{c}{0.488} \\
		TeCo      & \multicolumn{2}{c}{0.518} 	& \multicolumn{2}{c}{0.693} \\
		\bottomrule
	\end{tabular}
    \end{small}
%	\vspace{1em}
\end{table}

\textbf{Performance with Commonly-Used Open-Source BackdoorBench.} %Besides the evaluation on our self-defined platform with traditional BadNet as trigger-generating manner, 
We test more advanced backdoor attacks on the commonly-used BackdoorBench \cite{BackdoorBench} regrading ResNet-18 as target model. %such as the dynamic backdoor attack-Input-Aware \cite{InputAware} and imperceptible warping-based Backdoor Attack-WaNet \cite{WaNet}.%, to demonstrate the broad defensive adaptability of our SifterNet. The Input-Aware attack aims to break the traditional characteristics of BadNet attack, wherein the triggers are fixed in shape, color and position. Under Input-Aware attack, each poisoned image is embedded into a trigger with different shape, color and position, in this way, it can effectively bypass certain assumption and premise of defense countermeasures, thereby rendering the static defense ineffective. Differently, WaNet does not implant backdoor through adding additional image, instead, it uses a transformation method known as Thin-Plate Spline (TPS) to creates a smooth deformation field across the entire image. This transformation field is generated through uniformly distributed control points throughout the image, which are slightly and randomly shifted to create a smooth, imperceptible stretching effect. Alike Input-Aware, WaNet also avoids some defense assumptions and also bypasses human visual detection to a certain extent. 
We compare our method with the integrated STRIP \cite{STRIP} and TeCo \cite{TeCo} in BackdoorBench. They are both black-box defense methods as well, i.e., no requirement on details neural network. The core idea of TeCo is to add various interference and noise to the image to determine whether the trigger is contained or not. TeCo assumes that adding noise would impact the target model’s recognition of triggers more than that of normal semantic regions. Thus, if multiple types of noises are added to an image, forming a small test set, the prediction results for trigger-implanted images will often vary due to the disruption of trigger structure. Conversely, if the image is clean, the predictions will remain more consistent. The TeCo method uses the consistency of prediction results across the test set to judge whether the original image is clean or not.

Table \ref{tab:cifar10_results} shows the experimental results, wherein our SifterNet still uses the metrics $Acc.$ and $ASR$ due to the purification on images, but the baselines STRIP and TeCo take the metric $AUC$ (Area Under Curve) due to the detection on images\footnote{$AUC$ is usually used for the effectiveness evaluation on detection tasks}. From the experimental results, we have the following observations: i) our SifterNet decreases the $ASR$ by 82.03\% for BadNet, 49.91\% for Input-Aware, 78.96\% for WaNet, 44.53\% for Blended, and 46.04\% for SIG on CIFAR-10 dataset, while declines the $Acc.$ by 15\%, 20.92\%, 22.25\%, 14\%, and 8.52\% respectively. Compared to the similar method IFMV, our SifterNet has a remarkable reduction on $ASR$ while the $Acc.$ is close. Moreover, STRIP has a very low $AUC$ 0.417 under Input-Aware, 0.433 under WaNet, and 0.488 under SIG, which discloses the trigger-detection results under advanced backdoor attacks are much poorer than that under traditional BadNet attack. Compared to STRIP, TeCo behaves better with an obvious lift although the $AUCs$ under Blended and SIG are not up to 0.6 and 0.7. This set of results unveils that our method can still significantly degrade the rate of successful attack with a limited accuracy loss even under the advanced backdoor attacks; ii) compared to BadNet and WaNet, the advanced backdoor attack Input-Aware, Blended, and SIG are stronger, that is to say, our method enables the $ASR$ to degrade to no more than 14\% under BadNet and 5\% under WaNet, however, the $ASRs$ reach at 48.60\% under Input-Aware, 38.38\% under Blended, and 37.22\% under SIG on CIFAR-10; and iii) for the baselines, TeCo outperforms STRIP. STRIP itself achieves strong detection performance under BadNet, stemming from an elegant core assumption by design, i.e. when a poisoned image is overlaid with a clean image, the trigger should result in the designated label after the combined image is passed through the target model. However, STRIP inherently assumes the trigger is fixed and operates independently from the main semantic region of the image. Such assumption can significantly limit the generalizability, and this limitation is evident from the performance drop of STRIP when facing other attacks except for BadNet.  

In addition, TeCo executes backdoor detection through observing the difference in prediction consistency between poisoned and clean samples. Yet, this non-targeted attack requires to generate a large number of noisy images with different types and intensities to improve detection effectiveness. In our experiments, we applied 15 types of noise, each with five intensity levels, resulting in a test set of 75 images for each test image, which significantly increases the computational demand. The main drawback of this non-targeted backdoor attack is the high computation cost. Thus, compared to the lightweight time overhead of our SifterNet and STRIP, TeCo is time-consuming. The analogous experimental and analysis can be obtained on dataset GTSRB in Table \ref{tab:gtsrb_results} in Appendix \ref{Sec:openSourcePlatformExperiment}.

\textbf{Performance Using High-Resolution Dataset and Vision Large Model.} Until now, the datasets used in the experiments are low-resolution and target models belong to traditional convolution neural networks. We continuously explore the effectiveness on high-resolution ImageNet and vision large model ViT. The experiments and analytics are depicted in Appendix \ref{Sec:openSourcePlatformExperiment}, from which we can conclude that our SifteNet can still significantly reduce the $ASR$, although the $Acc.$ becomes sensitive to the high-resolution images and ViT. 

\textbf{Computation Complexity.} To verify the lightweight property of our SifterNet, we fun a set of experiments using our self-built experiment platform and the open-source BackdoorBench, the experimental results and analytics are demonstrated in Appendix \ref{Sec:timeOverhead}, from which we obtain the conclusion that our SifterNet can achieve a remarkable lightweight computation. 

\subsection{Robustness enhancement} \label{Sec: RobustEnhance}
Taking into consideration the side-effect of trigger purification more or less, we also figure out two fashions to enhance the routine accuracy from the perspectives of Hopfield-network expansion and data augmentation in Appendix \ref{Sec:robustEnhancement}.

\section {Related work and discussion}
We review the related work in Appendix \ref{Sec: RelatedWork}, and furtherly discuss the encountered limitations in the aspects of handling vision large models and large number of categories of datasets, and potential solutions by our SifterNet and other baselines in Appendix \ref{Sec: OpenIssue}. 

\section{Conclusion} \label{Sec: Conclusion}
Inspired by Ising Model in classical physics, this work proposes a novel Hopfield Network-based backdoor defense countermeasure through appropriately purifying the implanted triggers resorting the "Memorization-Association" functionality of Hopfield network. From the theoretical perspective, we also provide formal analysis to showcase the existence of stability with Hebbian Learning, which guarantees the trigger can be purified appropriately. Multi-facet experiments on five commonly-used low-resolution and high-resolution datasets validate the effectiveness and efficiency of our approach under five ordinary and advanced representative backdoor attacks. Compared to the state-of-the-art trigger-purification and trigger-detection methods, even with the vision large model ViT, our proposed SifterNet can achieve superior performance on trigger purification under the guidance of memorization-association functionality. Furthermore, the properties of lightweight, generalization, agnosticism of target model and attack behaviors also make our method particularly adaptable to deploy in practical environment. In addition, we believe our work will shed light on the deep learning-related research from a new perspective of classical physics.

%\section*{References}
%{
%	\small
%	\bibliographystyle{unsrtnat}    % 或者 plainnat
%	\bibliography{IEEEexampleFan}
%}

\bibliography{example_paper}
\bibliographystyle{neurips_2025}

%%%%%%%%%%%%%%%%%%%%%%%%%%%%%%%%%%%%%%%%%%%%%%%%%%%%%%%%%%%%

\appendix

%\section{Technical Appendices and Supplementary Material}
%Technical appendices with additional results, figures, graphs and proofs may be submitted with the paper submission before the full submission deadline (see above), or as a separate PDF in the ZIP file below before the supplementary material deadline. There is no page limit for the technical appendices.

\section{Category of backdoor defense work} \label{Sec:categorybackdoordefense}
The state-of-the-art work on backdoor defense is categorized as shown in Table \ref{tab:backdoor-defense-methods}, in addition to our proposed SiferNet. 
\begin{table}[bthp]
	\caption{Category of backdoor-defense methods}
	\label{tab:backdoor-defense-methods}
	\begin{small}
	\centering
	%\vspace{1em}
	\begin{tabular}{lcccccccc}
		\toprule
		\textbf{Methods} & \textbf{T-LW} & \textbf{T-CI} & \textbf{T-WB} & \textbf{T-BB} & \textbf{I-LW} & \textbf{I-CI} & \textbf{I-WB} & \textbf{I-BB} \\
		\midrule
		SifterNet(Ours)           &      &      &      &      & $\surd$ &      &      & $\surd$ \\
		IFMV \cite{IFMV}          &      &      &      &      & $\surd$ &      &      & $\surd$ \\
		STRIP \cite{STRIP}        &      &      &      &      & $\surd$ &      &      & $\surd$ \\
		TECO \cite{TeCo}          &      &      &      &      &        & $\surd$ &      & $\surd$ \\
		Sentinet \cite{SentiNet}  &      &      &      &      &        & $\surd$ & $\surd$ &      \\
		BEATRIX \cite{BEATRIX}    &      & $\surd$ & $\surd$ &      &        &        &        &      \\
		SCAN \cite{Demon_in_the_Variant} &      & $\surd$ & $\surd$ &      &        &        &        &      \\
		SPECTRE \cite{SPECTRE}    &      & $\surd$ & $\surd$ &      &        &        &        &      \\
		SS \cite{SS}              &      & $\surd$ & $\surd$ &      &        &        &        &      \\
		AGPD \cite{AGPD}          &      & $\surd$ & $\surd$ &      &        &        &        &      \\
		\bottomrule
		\multicolumn{9}{l}{\makecell{T: Training Stage \quad I: Inference Stage \quad LW: Light-Weight \quad CI: Computation-Intensive \quad \\ WB: White-Box \quad BB: Black-Box}} \\
	\end{tabular}
	%\vspace{1em}
	\end{small}
\end{table}

\section{Stability analysis on hopfield network} \label{stabilityHebbian}
\textbf{Hopfield Network Setup.}
Consider a fully connected, symmetric Hopfield network with $N$ binary neurons
$x_i\in\{+1,-1\}$.
The energy of a state $x=(x_1,\dots,x_N)$ is
\begin{equation}
	E(x)=-\frac12\sum_{i\neq j}w_{ij}x_i x_j,
\end{equation}
where $w_{ij}=w_{ji}$ and $w_{ii}=0$.
Asynchronous single-spin updates $x_i\!\leftarrow\!\mathrm{sgn}(h_i)$ with local field $h_i=\sum_{j\neq i}w_{ij}x_j$ always satisfy $E(\text{new})\le E(\text{old})$, so $E$ is a Lyapunov function and the dynamics must converge to a local minimum.

\textbf{Hebbian Weights.}
Suppose we want to store $P$ random patterns $\xi^{\mu}\in\{+1,-1\}^{N}$ ($\mu=1,\dots,P$). Hebbian learning sets
\begin{equation}
	w_{ij}= \frac1N\sum_{\mu=1}^{P}\xi_i^{\mu}\xi_j^{\mu},
	\qquad i\neq j. \label{eq:hebb}
\end{equation}
The $1/N$ factor keeps the local fields $\mathcal{O}(1)$ as $N\to\infty$.

\textbf{Local‑Field Criterion for Stability.}
A pattern $\xi^{\mu}$ is a strict local minimum if every neuron aligns with its own local field:
\begin{equation}
	\xi_i^{\mu}h_i(\xi^{\mu})>0
	\quad\forall\,i. \label{eq:stability}
\end{equation}

Substitute Formula \ref{eq:stability} into Formula \eqref{eq:hebb}, we have that
\begin{align}
	h_i(\xi^{\mu})
	&=\sum_{j\neq i}\frac1N\sum_{\nu=1}^{P}\xi_i^{\nu}\xi_j^{\nu}\xi_j^{\mu}
	\notag\\
	&=\Bigl(1-\frac1N\Bigr)\xi_i^{\mu}
	+\underbrace{\frac1N\sum_{\nu\neq\mu}\sum_{j\neq i}
		\xi_i^{\nu}\xi_j^{\nu}\xi_j^{\mu}}_{\displaystyle\eta_i}. \label{eq:field}
\end{align}

The first term is deterministic, and the interference term $\eta_i$ is a sum of $(P-1)$ independent $\pm1$ variables, by the central-limit theorem, we have that
\begin{equation}
	\eta_i \sim \mathcal{N}\!\bigl(0,\tfrac{P-1}{N}\bigr).    \label{eq:noise}
\end{equation}

\textbf{Capacity Bound.} Condition~\eqref{eq:stability} holds with high probability if $\lvert\eta_i\rvert<1$ for all $i$.
Because $\eta_i$ has standard deviation $\sqrt{(P-1)/N}$, a union bound yields
\begin{equation}
	\Pr\bigl[\text{any spin flips}\bigr]
	\;\le\;N \,\exp\!\Bigl(-\frac{1}{2}\frac{N}{P-1}\Bigr).
\end{equation}

Thus, the failure probability vanishes as $N\!\to\!\infty$, provided
\begin{equation}
	\alpha\;=\;P/N < \alpha_c \approx 0.138,
\end{equation}
the celebrated storage capacity of the Hopfield model.

\textbf{Energy Difference.}
Write the energy difference between a stored pattern and a one-bit flip: choose neuron $k$, define $x=\xi^{\mu}$ and $x'$
identical except $x'_k=-\xi_k^{\mu}$. Because Formula \eqref{eq:stability} ensures $\xi_k^{\mu}h_k>0$, 
\begin{equation}
	E(x')-E(x)=2\xi_k^{\mu}h_k>0.
\end{equation}
$E$ strictly increases along every single-spin departure from $\xi^{\mu}$, consequently, $\xi^{\mu}$ is a local minimum.

\textbf{Convergence of Corrupted Inputs.}
Any initial pattern $x^{(0)}$ evolves by monotonically decreasing $E$. If its Hamming distance to some stored pattern $\xi^{\mu}$ is below the basin radius (order $O(N)$ for $\alpha<\alpha_c$), the trajectory cannot cross an energy barrier
and must fall into $\xi^{\mu}$. Otherwise, it may land in a spurious minimum created by the $\eta_i$ noise.

Hebbian learning embeds each training pattern as a strict local minimum of the Hopfield energy landscape, guaranteed whenever the load factor $\alpha=P/N$ is below the critical capacity $0.138$. The Lyapunov property of $E$ then drives any sufficiently corrupted version of a stored pattern back to the original, explaining the associative recall in Hopfield network.

\section{Algorithms sketch} \label{algorithm}
\begin{algorithm}[tbhp]
	\caption{Hopfield Network Training}
	\label{Algorithm-1}
	\begin{algorithmic}
		\REQUIRE \texttt{testset\_clean}, a dataset of test images
		\ENSURE \texttt{weight}, the final trained weight matrix
		\STATE Initialize an empty list \texttt{put\_array} for storing processed data
		\STATE Initialize \texttt{weight} as a zero matrix of size (array\_length, array\_length)
		\STATE Initialize \texttt{index} as 0
		\FOR{each image batch in \texttt{testset\_clean}}
		\STATE Convert each image in the batch to binary values based on a threshold
		\STATE Flatten the binary image to a one-dimensional array
		\STATE Store the flattened array in \texttt{put\_array}
		\IF{end of designated batches}
		\STATE Break the loop
		\ENDIF
		\STATE Increment \texttt{index}
		\ENDFOR
		\FOR{each stored data in \texttt{put\_array}}
		\STATE Initialize a temporary zero matrix \texttt{W0} for pairwise weights
		\STATE Calculate the weights using pairwise multiplication of the data elements
		\STATE Update the \texttt{weight} matrix by adding \texttt{W0}
		\ENDFOR
		\STATE \textbf{Return} \texttt{weight}
	\end{algorithmic}
\end{algorithm}

\begin{algorithm}[tbhp]
	\caption{Hopfield Network Processing}
	\label{Algorithm-2}
	\begin{algorithmic}
		\REQUIRE \texttt{img}, an unknown input image
		\ENSURE Processed \texttt{img}
		\STATE Convert \texttt{img} to binary values based on a threshold
		\STATE Flatten the binary image to a one-dimensional vector \texttt{V0}
		\FOR{a set number of iterations}
		\STATE Randomly select an index \texttt{i}
		\STATE Calculate the input \texttt{u} using the weight matrix and \texttt{V0}
		\STATE Update the value at index \texttt{i} in \texttt{V0} based on \texttt{u}
		\ENDFOR
		\STATE Reshape \texttt{V0} back to the original image dimensions and update \texttt{img}
		\STATE \textbf{Return} \texttt{img}
	\end{algorithmic}
\end{algorithm}

\section{Experiment \& performance}
\subsection{Configuration} \label{Sec:Configuration}
\textbf{Datasets.} Four commonly-used datasets are used to evaluate the effectiveness and efficiency of our approach: MNIST \cite{LeCun1998}, FashionMNIST \cite{Fashion_MNIST}, CIFAR-10 \cite{Krizhevsky2009} and GTSRB \cite{Houben13}, and Table \ref{Table:dataset} presents the statistics information. 

\begin{table}[bth] 
\caption{Dataset statistics.}
\label{Table:dataset}
\begin{small}
\centering
\newcolumntype{C}[1]{>{\centering\let\newline\\\arraybackslash\hspace{0pt}}m{#1}}
\begin{tabular}{C{0.12\textwidth} C{0.15\textwidth} C{0.08\textwidth} C{0.12\textwidth} C{0.08\textwidth} C{0.26\textwidth}}
	%	\begin{tabular}{cccccc}
		\toprule
		\textbf{Dataset} & \textbf{Parameter} & \textbf{Label} & \textbf{Image Size} & \textbf{Image} & \textbf{Model Architecture}  \\
		\midrule
		MNIST & 429,290 & 10 & 28$\times$28$\times$1 & 60,000 & 2Conv. + 2Dense + 1Dropout \\
		CIFAR-10 & 429,290 & 10 & 32$\times$32$\times$3 & 60,000 & 2Conv. + 2Dense + 1Dropout \\
		FashionMNIST & 429,290 & 10 & 28$\times$28$\times$1 & 60,000 & 2Conv. + 2Dense + 1Dropout \\
		GTSRB & 2,017,131 & 43 & 32$\times$32$\times$3 & 50,000 & 5(Conv. + Pool) + 1Conv. + 1Flatten + 2Dense \\
		\bottomrule
\end{tabular}
\end{small}
\end{table}

\textbf{Baselines.} 
To sufficiently verify the effectiveness of our method, the following four representative backdoor defense methods are employed:
\begin{itemize} 
	\item \textbf{STRIP \cite{STRIP}.} This work intends to recognize the trigger’s pattern, regardless of the main semantic area of the contaminated image. The main idea is at first to choose a set of clean dataset with sample size $n$, then overlap the input sample with the clean samples. Subsequently, input these $n$ mixed samples to target model for task inference, and observe the sum of Shannon entropy of the prediction results. If the results are well-dispersed, it indicates the input image is clean. Oppositely, if the outputs are concentrated, which hints the image contains a trigger.
	\item \textbf{SentiNet \cite{SentiNet}.} This mitigation aims to locate the trigger through resorting to the saliency mapping between trigger-implanted and clean samples. For suspicious areas, they are sequentially pasted onto images from a clean dataset. Comparatively, it also pasts inert Gaussian noise onto the same clean dataset. If the images pasted by suspicious area make the prediction results consistent, then the area is judged as a trigger.
	\item \textbf{Input Fuzzing and Majority Voting (IFMV) \cite{IFMV}.} IFMV concentrates on eliminating trigger by adding random (Gaussian/local) noise. However, the challenge lies in how to add adequate noise and simultaneously maintain the core-semantics invariant. %This method is similar as ours, however, it is unstable during trigger elimination deriving from inadequately controlling on the amount of noise. 
	\item \textbf{TeCo \cite{TeCo}.} Compared to the IFMV approach, which attempts to remove backdoors by simply injecting noise, the TeCo method instead focuses on the robustness gap between trigger-embedded images and clean images. It repeatedly adds noise of varying strengths to the input under test and monitors how consistently the model’s predictions hold across these noisy versions. If a sample’s prediction remains highly consistent despite different noise levels, it is likely a trigger-carrying adversarial image. However, achieving reliable detection this way typically requires a large variety of noise types and the generation of multiple noise intensities for each type—resulting in a substantial time overhead. 
\end{itemize}

\textbf{Backdoor Attacks.} To comprehensively verify the defense capability of our work, the following five routine backdoor attacks in our experiments are employed to generate different types of triggers.
\begin{itemize}
	\item \textbf{BadNet \cite{BadNets}.} The target of BadNet attack is class 0. The model used in the experiment is ResNet18, with a total of 100 training epochs. The optimizer used is SGD with a momentum of 0.9. The learning rate is set to 0.01, and the CosineAnnealingLR scheduler is used for learning rate scheduling. The attack employs the all2one label transformation strategy, which converts all input image labels to the target label (class 0). During training, a 3x3 white patch is added as the trigger, placed at the bottom-right corner of the image. 10\% of the training images will be injected with this trigger (pratio=0.1).
	\item \textbf{Input-Aware \cite{InputAware}.}  The Input-Aware dynamic backdoor attack targets class 0. The model used in the experiments is ResNet18, with input image dimensions of 32x32x3, and a total of 100 training epochs. The optimizer used is SGD, with a momentum of 0.9 and a learning rate of 0.01. The cosine annealing learning rate scheduler (CosineAnnealingLR) is also applied. The attack adopts the all2one label transformation strategy, where all input images' labels are converted to the target label (class 0). During training, image augmentation techniques such as random cropping (RandomCrop=5) and random rotation (RandomRotation=10) are applied. To ensure the diversity and non-reusability of the triggers, diversity loss (LambdaDiv=1) is used to enforce significant difference between the generated triggers for different inputs. Additionally, regularization loss (LambdaNorm=100) is applied to constrain the shape of the triggers and prevent from overfitting.
	\item \textbf{WaNet \cite{WaNet}.}  The target of WaNet attack is class 0. The model used in the experiments is ResNet18, with input image dimensions of 32x32x3, and a total of 100 training epochs. The optimizer used is SGD, with a momentum of 0.9 and a learning rate of 0.01. The MultiStepLR learning rate scheduler is applied, with the learning rate decreasing by a factor of 10 at the 100th, 200th, 300th, and 400th epochs. The attack adopts the all2one label transformation strategy, where all input image labels are converted to the target label (class 0). During training, 10\% of the images will be injected with attack triggers (pratio=0.1). To generate the triggers, a 4x4 grid (k=4) is used to define the deformed region, with a deformation strength 0.5 (s=0.5). To enhance the stealthiness of the attack, the experiments do not use random cropping (RandomCrop=0) or random rotation (RandomRotation=0) for image augmentation.	
	\item \textbf{Blended \cite{Blended}.} In this experiment, we perform classification and a backdoor attack on ResNet-18 using a batch size of 128, random seed 0, and four data-loading workers, with pin\_memory and non\_blocking enabled to boost I/O performance (prefetch and mixed precision disabled). The optimizer is SGD (momentum 0.9, weight decay 0.0005) with an initial learning rate of 0.01, and we train for 100 epochs using a CosineAnnealingLR scheduler, without saving intermediate checkpoints. The backdoor attack is of the blended type: the image “hello\_kitty.jpeg” is used as the trigger, blended at an alpha of 0.065 during both training and testing; the attack label transformation is all-to-one targeting class 0, with an injection ratio of 7.5%.
	\item \textbf{SIG \cite{SIG}.} We employ a signal-triggered backdoor attack (attack: sig) with signal amplitude delta = 5 (sig\_delta: 5) and frequency 6 (sig\_f: 6), mapping all samples to the target label 0 (attack\_label\_trans: all2one, attack\_target: 0) and poisoning 7.5\% of the training set (pratio: 0.075). The batch size is 128 (batch\_size: 128), and we optimize with SGD using momentum 0.9 (client\_optimizer: sgd, sgd\_momentum: 0.9, wd: 0.0005). The initial learning rate is 0.01 (lr: 0.01), scheduled by CosineAnnealingLR (lr\_scheduler: CosineAnnealingLR). To ensure reproducibility, the random seed is fixed at 0 (random\_seed: 0), and training runs for 100 epochs (epochs: 100).
\end{itemize}

\textbf{Metrics.} Two regular and common performance metrics are provided: Accuracy ($Acc.$) and attack success rate ($ASR$). The former reflects the inference accuracy on the purified samples from whatever benign or trigger-implanted images; and the latter denotes the rate of successfully launching backdoor attacks towards target model.

\textbf{Execution Environment.} All the experiments are performed in the following environment, i.e. CPU: Intel core i7 10750H; GPU: NVIDIA GeForce RTX 2070 SUPER Max-Q; Internal Storage: 16GB. Target Model Training Time: 2$\sim $4 hours; Hopfield Network Training Time: 30 $\sim$120 sec.

\subsection{Memorization iteration setting} \label{memorizationIterationSetting}
\begin{figure*} [hbtp]
	\centering
	\subfigure[MNIST]{
		\begin{minipage}[t]{0.455\columnwidth}
			\centering
			\includegraphics[width=2.6in, height=1.2in]{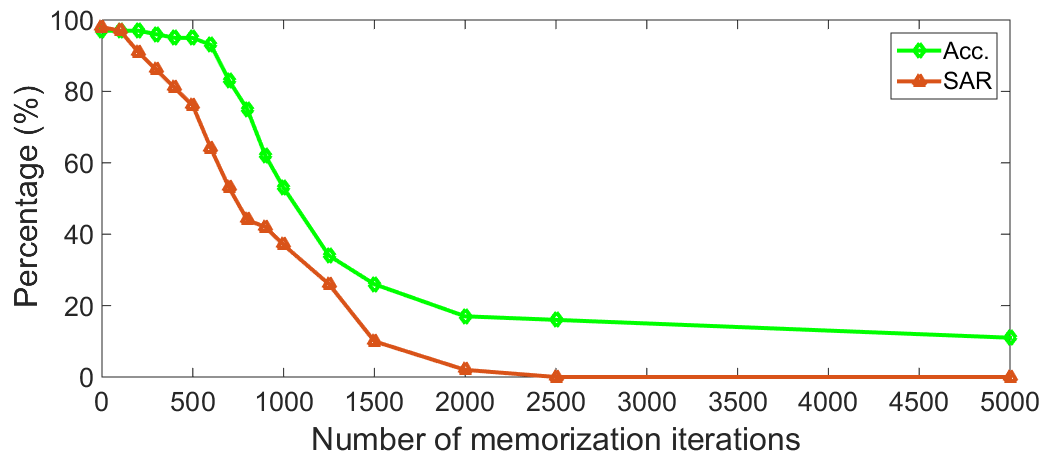}
			%	\caption{Rank growth under adversarial attacks}
			\label{fig:AccCurve_MNIST}
		\end{minipage}
	}
	\vspace{0.02cm}
	%     \quad
	\subfigure[CIFAR-10]{
		\begin{minipage}[t]{0.455\columnwidth}
			\centering
			\includegraphics[width=2.6in, height=1.2in]{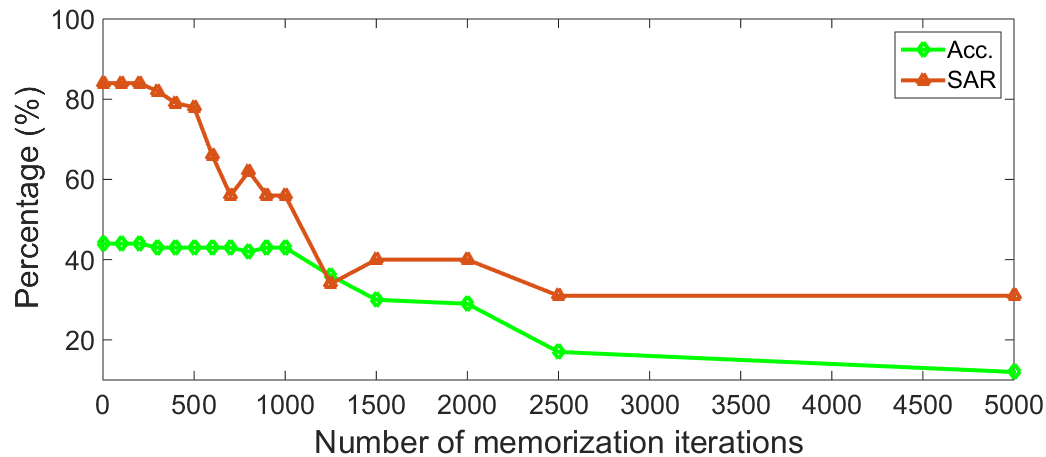}
			%	\caption{Rank decrease with our EquiliRes}
			\label{fig:AccCurve_CIFAR10} 
		\end{minipage}
	}
	\vspace{0.02cm}
	%     \quad
	\subfigure[FashionMNIST]{
		\begin{minipage}[t]{0.455\columnwidth}
			\centering
			\includegraphics[width=2.6in, height=1.2in]{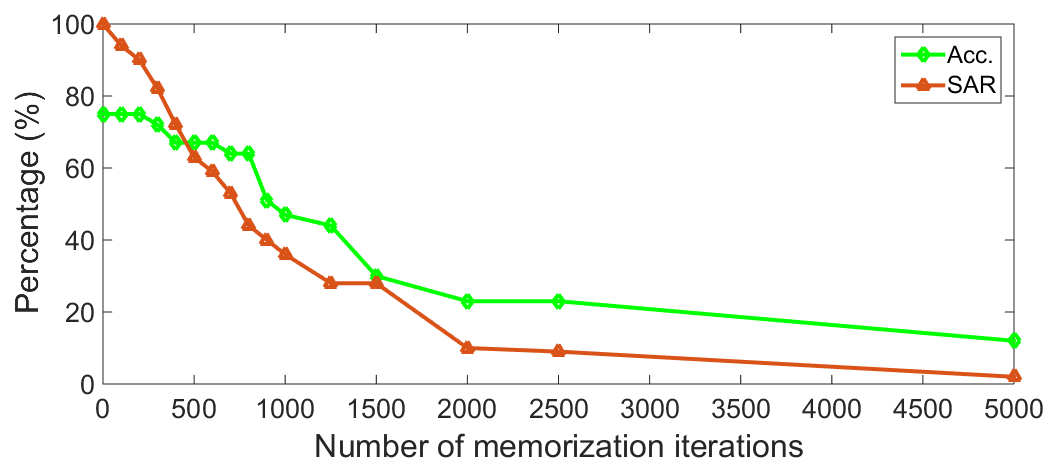}
			%	\caption{Rank decrease with our EquiliRes}
			\label{fig:SARCurve_FashionMNIST} 
		\end{minipage}
	}
	\vspace{0.02cm}
	%     \quad
	\subfigure[GTSRB]{
		\begin{minipage}[t]{0.455\columnwidth}
			\centering
			\includegraphics[width=2.6in, height=1.2in]{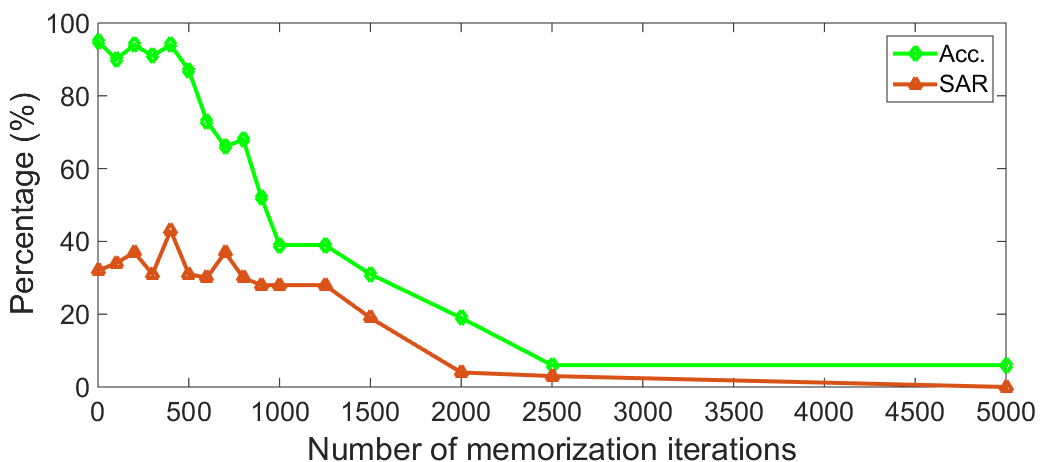}
			%	\caption{Rank decrease with our EquiliRes}
			\label{fig:SARCurve_GTSRB} 
		\end{minipage}
	}
	\caption{The $Acc.$ and $ASR$ performance as the number of memorization iterations increases.}
	\label{fig:AccandASR}
\end{figure*}

Our proposed SifterNet employs the memorization functionality in an iterative means, the number of iterations also has a suitable choice, in other words, too few iterations may lead to weak trigger-purification, while too many may excessively alter the original shape, leading to the reduction of target-model accuracy. Seen from Fig. \ref{fig:AccandASR} , the two metrics $ASR$ and $Acc.$ both decrease as the number of memorization iterations enlarges on the four datesets, that is to say, $ASR$ anticipates more iterations but $Acc.$ does not. Pursuant to the performance curves, we know that $Acc.$ has a relatively steep decrease with each dataset. Therefore, we locate the number at the steep point when the accuracy begins to deteriorate as the selected number of memorization. 

Seen from the experimental results, the reasonable ranges of memorization iterations are respectively in [800, 1200] for MNIST, [900, 1200] for CIFAR-10, [800, 1000] for FashionMNIST, and [1, 500] for GTSRB, since the purification has somewhat randomness, the optimal purification can be accomplished within the intervals. 

In backdoorbench, the open source test platform, we all use three-channel SifterNet defenses based on local differentiation. The main parameters involved include the radius $k$-size of surrounding pixels considered when calculating whether each pixel is lit in local differentiation, and the number of Hopfield random convergence RemoveTime. In CIFAR-10 dataset, $k$-size is generally the best in [15, 25], and RemoveTime is the best in [50, 500]. In GTSRB dataset, $k$-size is generally best in [5, 15], and RemoveTime is best in [50, 500].
%Fig. \ref{fig:smallTrigger} and Fig. \ref{fig:largeTrigger} exhibit the trigger-purifying statuses under the defense approach IFMV [?].

\subsection{Volume of clean samples vs. purification effectiveness} \label{Sec:cleanSamplevsPurification}
Hopfield network training requires few clean samples, we here give an investigation on requirement of the clean sample quantity, regarding the correlation with purification effectiveness. The statistics is listed in Table \ref{table:CleanSampleNum}.

\begin{table}[ht]
	
	\caption{Number of clean samples and the defense effect}
	\label{table:CleanSampleNum}
	
	\centering
	\begin{tabular}{ccc}
		\toprule
		clean sample & Acc.    & ASR     \\
		\midrule
		64           & 56.51\% & 6.25\%  \\
		48           & 60.00\% & 6.25\%  \\
		32           & 61.29\% & 3.12\%  \\
		16           & 60.00\% & 6.25\%  \\
		8            & 50.00\% & 12.50\% \\
		4            & 48.27\% & 9.38\%  \\
		\bottomrule
	\end{tabular}
	
\end{table}

\subsection{Performance with our self-built neural network model} \label{Sec:selfBuiltPlatformExperiment}
Upon the four datasets, we implements a set of experiments and obtain the results as shown in Table \ref{table:performance-tca-asr} and Table \ref{table:ifmv-performance}. The former shows the performance of our SifterNet, from which we can observe that our proposed approach has excellent performance across whatever single-channel or three-channel images, namely the $ASR$ declines by 59.04\%, 60.98\%, 55.15\% and 65.62\% with the datasets MNIST, CIFAR-10, FashionMNIST and GTSRB respectively, while the $Acc.$ decreases by 17.77\%, 7.69\%, 12.73\% and 7.00\% respectively. These experimental results bring two hints: i) the purification on trigger indeed accompanies side-effect on routine accuracy; and ii) our approach indeed enables to gain a large extent of reduction of successful backdoor attack with little accuracy sacrifice, which further indicates the seed-dateset-based Hopfield network training can behave well in both aspects of memorization and association, resulting in appropriate trigger purification. 

\begin{table}[tbhp]
\centering
\caption{The Performance of Our SifterNet (Trigger: BadNet; Neural Network: set in Table \ref{Table:dataset})}
\label{table:performance-tca-asr}
%	\vspace{1em}
\begin{small}
%	\begin{tabular}{lcccc}  % 使用标准的列格式
\newcolumntype{C}[1]{>{\centering\let\newline\\\arraybackslash\hspace{0pt}}m{#1}}
\begin{tabular}{C{0.108\textwidth} C{0.078\textwidth} C{0.095\textwidth} C{0.295\textwidth} C{0.295\textwidth}}	
\toprule  % 表格上方的线
\textbf{Dataset} & \multicolumn{2}{c}{\textbf{Before SifterNet}} & \multicolumn{2}{c}{\textbf{After SifterNet}} \\ \midrule  % 中间的横线
& \textbf{Acc.} & \textbf{ASR} & \textbf{Acc.} & \textbf{ASR} \\ \midrule  % 中间的横线
MNIST & 97.80\% & 99.80\% & 80.03\%$\pm$13.70\% (17.77\% $\downarrow$) & 40.76\%$\pm$12.50\% (59.04\% $\downarrow$) \\
CIFAR-10 & 43.75\% & 100.00\% & 36.06\%$\pm$6.50\% (7.69\% $\downarrow$) & 39.02\%$\pm$6.00\% (60.98\% $\downarrow$) \\
FashionMNIST & 75.00\% & 98.90\% & 62.27\%$\pm$6.50\% (12.73\% $\downarrow$) & 43.75\%$\pm$10.00\% (55.15\% $\downarrow$) \\
GTSRB & 98.75\% & 100.00\% & 91.75\%$\pm$5.00\% (7.00\% $\downarrow$) & 34.38\%$\pm$5.00\% (65.62\% $\downarrow$) \\
\bottomrule  % 表格下方的线
\end{tabular}
%	\vspace{1em}
\end{small}
\end{table}

\begin{table}[tbhp]
\centering
\caption{The Performance of IFMV (Trigger: BadNet; Neural Network: set in Table \ref{Table:dataset})}
\label{table:ifmv-performance}
\begin{small}
%	\vspace{1em}
%	\begin{tabular}{lcccc}  % 使用标准的列格式
\newcolumntype{C}[1]{>{\centering\let\newline\\\arraybackslash\hspace{0pt}}m{#1}}
\begin{tabular}{C{0.11\textwidth} C{0.08\textwidth} C{0.1\textwidth} C{0.28\textwidth} C{0.28\textwidth}}	
\toprule  % 表格上方的线
\textbf{Voting} & \multicolumn{2}{c}{\textbf{Before Voting}} & \multicolumn{2}{c}{\textbf{After Voting}} \\ \midrule  % 中间的横线
& \textbf{Acc.} & \textbf{ASR} & \textbf{Acc.} & \textbf{ASR} \\ \midrule  % 中间的横线
MNIST & 97.80\% & 99.90\% & 71.42\%$\pm$9.70\% (26.38\%$\downarrow$) & 89.06\%$\pm$5.60\% (10.74\%$\downarrow$) \\
CIFAR-10 & 43.75\% & 100.00\% & 25.00\%$\pm$5.00\% (18.75\%$\downarrow$) & 90.75\%$\pm$6.00\% (9.25\%$\downarrow$) \\
FashionMNIST & 75.00\% & 98.90\% & 50.00\%$\pm$5.00\% (25.00\%$\downarrow$) & 80.00\%$\pm$5.00\% (18.90\%$\downarrow$) \\
GTSRB & 98.75\% & 100.00\% & 57.90\%$\pm$5.60\% (40.85\%$\downarrow$) & 78.12\%$\pm$7.70\% (21.88\%$\downarrow$) \\
\bottomrule  % 表格下方的线
\end{tabular}
%	\vspace{1em}
\end{small}
\end{table}

\begin{figure} [tbhp]
\centering
\subfigure[Small-trigger elimination]{
%	\subfloat[2D Ising model with lattice arrangement]{
	\begin{minipage}[t]{0.35\columnwidth}
		\centering
		\includegraphics[width=1.2in, height=1.1in]{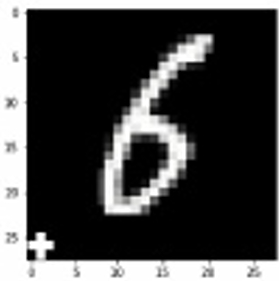}
		%	\caption{xx}
		\label{fig:largeTrigger}
	\end{minipage}
}
\vspace{0.02cm}
%     \quad
\subfigure[Large-trigger elimination]{
	%	\subfloat[Black-white image represented by magnetic pins]{
		\begin{minipage}[t]{0.35\columnwidth}
			\centering
			\includegraphics[width=1.2in, height=1.1in]{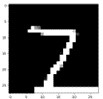}
			%	\caption{xx}
			\label{fig:Small-LargeTrigger} 
		\end{minipage}
	}
	\caption{Trigger purification on MNIST images.}
	\label{fig:Small-LargeTrigger}
\end{figure}

Table \ref{table:ifmv-performance} shows the experimental results of IFMV, a similar method with ours, from which we can observe the $ASR$ declines by 10.74\%, 9.25\%, 18.90\%, 21.88\% on datasets MNIST, CIFAR-10, FashionMNIST and GTSRB accompanying the $Acc.$ declining by 26.38\%, 18.75\%, 25.00\%, 40.85\% respectively. As aforementioned, it is data-dependent, the performance would deteriorate obviously when the datasets are changed. For example, we use MNIST datasets to test its stability and adaptability. The experimental results are shown in Fig. \ref{fig:Small-LargeTrigger}, which indicates IFMV can successfully eliminate 90\% of the triggers, %and 45\% for CIFAR-10, 
exhibiting a good performance. Nevertheless, when the trigger is enlarged to 20$\times$2 as shown in Fig. \ref{fig:largeTrigger}, it becomes deteriorated dramatically. That is to say, it would work well when an explicit disparity exists between the main-semantics area and trigger size, this is because the added noise can easily break up the small trigger to preserve the larger main-semantics area. However, as the trigger-size enlarges, it becomes difficult to keep such balance. For an in-depth analysis, we think IFMV cannot recover the contaminated sample to clean state in a proper manner, unlike our approach that can achieve proper purification under the guidance of memorization and association functionalities via the "attractor domains".
%\begin{figure}[htbp]
%	\centering
%	\begin{minipage}[t]{0.45\columnwidth}
%		\centering
%		\includegraphics[height=1.2in]{smallTrigger.png}
%		\caption{Small-trigger elimination.}
%		\label{fig:smallTrigger}
%	\end{minipage}
%	\hfill
%	\begin{minipage}[t]{0.45\columnwidth}
%		\centering
%		\includegraphics[height=1.2in]{largeTrigger.png}
%		\caption{Large-trigger elimination.}
%		\label{fig:largeTrigger}
%	\end{minipage}
%\end{figure}

%\begin{figure}[ht]
%\centering
%
%\subfigure[Small-trigger elimination]{
%\includegraphics[width=0.35\columnwidth]{smallTrigger.png}
%\label{fig:smallTrigger}
%}
%\hfill
%\subfigure[Large-trigger elimination]{
%\includegraphics[width=0.35\columnwidth]{largeTrigger.png}
%\label{fig:largeTrigger}
%}
%
%\caption{Trigger purification on MNIST images.}
%\label{fig:Small-LargeTrigger}
%\end{figure}

In addition, we also perform a set of experiments to evaluate STRIP \cite{STRIP} and SentiNet \cite{SentiNet}, both of which aim to identify triggers rather than eliminating. STRIP performs well both in single-channel and three-channel images. It can achieves over 95\% detection on contaminated samples with MNIST and GTSRB datasets, nevertheless, the severe weakness lies in that it is hard to handle those active backdoor attacks, for instance, STRIP can only identify 10\%-15\% compromised inputs, deriving from the fact that active backdoor attacks can disrupt the characteristics of passive backdoor attacks. Similarly, SentiNet can also detect over 90\% trigger-compromised inputs with MNIST and GTSRB datasets. Given the detection is executed by pasting suspicious areas onto clean samples, thus the active backdoor attack would influence detection effectiveness as well, this is because such kind of active backdoor attacks not only relies on trigger, but it is also input-oriented, that is to say, for most of clean samples, although pasted onto triggers already, they may not activate the target model. In this case, it can merely identify less than 15\% contaminated inputs. %The study [3] also presents that active backdoor attack can degrade the capability of SentiNet to locate the trigger area. 

\subsection{Performance with open-source benchmark} \label{Sec:openSourcePlatformExperiment}

\begin{table}[tbhp]
\centering
\caption{Performance on GTSRB Datasets with BackdoorBench}
\label{tab:gtsrb_results}
\begin{small}
%	\vspace{1em}
\begin{tabular}{lcccccc}
%	\begin{tabular}{p{1.1cm}>{\centering\arraybackslash}p{1.32cm}>{\centering\arraybackslash}p{1.32cm}>{\centering\arraybackslash}p{1.32cm}>{\centering\arraybackslash}p{1.32cm}>{\centering\arraybackslash}p{1.32cm}>{\centering\arraybackslash}p{1.32cm}}	
	\toprule
	\multirow{2}{*}{\textbf{Defense}} &
	\multicolumn{2}{c}{\textbf{BadNet}} &
	\multicolumn{2}{c}{\textbf{Input-Aware}} &
	\multicolumn{2}{c}{\textbf{WaNet}} \\
	\cmidrule(lr){2-3}\cmidrule(lr){4-5}\cmidrule(lr){6-7}
	& Acc. & ASR & Acc. & ASR & Acc. & ASR \\
	\midrule
	Clean     & 88.37\% & 94.45\% & 88.28\% & 90.19\% & 58.24\% & 96.96\% \\
	SifterNet & 68.10\%$\pm$6\% & 44.56\%$\pm$10\% & 61.10\%$\pm$6.5\% & 43.75\%$\pm$5.5\% & 57.87\%$\pm$0.3\% & 0.03\%$\pm$0.01\% \\
	IFMV      & 58.48\% & 94.39\% & 37.54\% & 43.56\% & 18.94\% & 0.02\% \\
	\midrule
	& \multicolumn{2}{c}{\textbf{AUC}} & \multicolumn{2}{c}{\textbf{AUC}} & \multicolumn{2}{c}{\textbf{AUC}} \\
	\midrule  % 中间的横线
	STRIP     & \multicolumn{2}{c}{0.890}     & \multicolumn{2}{c}{0.514}   & \multicolumn{2}{c}{0.541} \\
	TeCo      & \multicolumn{2}{c}{0.931}     & \multicolumn{2}{c}{0.571}   & \multicolumn{2}{c}{0.801} \\
	\bottomrule
\end{tabular}

%	\vspace{1em}

%—— 第二部分（后半截列） ——
\begin{tabular}{lcccc}
	%	\begin{tabular}{p{1.1cm}>{\centering\arraybackslash}p{1.32cm}>{\centering\arraybackslash}p{1.32cm}>{\centering\arraybackslash}p{1.32cm}>{\centering\arraybackslash}p{1.32cm}}	
		\toprule
		\multirow{2}{*}{\textbf{Defense}} &
		\multicolumn{2}{c}{\textbf{Blended}} &
		\multicolumn{2}{c}{\textbf{SIG}} \\
		\cmidrule(lr){2-3}\cmidrule(lr){4-5}
		& Acc. & ASR & Acc. & ASR \\
		\midrule
		Clean     & 86.69\% & 84.41\% & 87.68\% & 85.77\% \\
		SifterNet & 71.23\%$\pm$4\% & 49.75\%$\pm$7\% & 62.63\%$\pm$3\% & 35.37\%$\pm$1\% \\
		IFMV      &  68.05\% & 61.13\% 	& 46.65\% & 85.06\% \\
		\midrule
		& \multicolumn{2}{c}{\textbf{AUC}} & \multicolumn{2}{c}{\textbf{AUC}} \\
		\midrule  % 中间的横线
		STRIP     & \multicolumn{2}{c}{0.578} & \multicolumn{2}{c}{0.465} \\
		TeCo      & \multicolumn{2}{c}{0.559} & \multicolumn{2}{c}{0.744} \\
		\bottomrule
	\end{tabular}
\end{small}
%	\vspace{1em}
\end{table}

\begin{table}[tbhp]
\centering
\caption{Performance on Tiny Imagenet Datasets with BackdoorBench}
\label{tab:Tiny_Imagenet}
\begin{small}
	\vspace{1em}
	\makebox[\linewidth][c]{%
		\begin{tabular}{lcccccccc}
			\toprule
			& \multicolumn{2}{c}{\textbf{BadNet}} 
			& \multicolumn{2}{c}{\textbf{Blended}} 
			& \multicolumn{2}{c}{\textbf{SIG}} 
			& \multicolumn{2}{c}{\textbf{WaNet}} \\
			\cmidrule(lr){2-3}\cmidrule(lr){4-5}\cmidrule(lr){6-7}\cmidrule(lr){8-9}
			\textbf{Defense / Detector} 
			& \textbf{Acc.} & \textbf{ASR} 
			& \textbf{Acc.} & \textbf{ASR} 
			& \textbf{Acc.} & \textbf{ASR} 
			& \textbf{Acc.} & \textbf{ASR} \\
			\midrule
			Clean      & 31.38\% & 87.58\% & 29.83\% & 90.83\% & 29.74\% & 92.19\% & 33.62\% & 96.99\% \\
			SifterNet  & 12.61\% & 16.11\% & 13.15\% & 45.93\% & 11.50\% & 27.86\% & 12.17\% &  1.63\% \\
			IFMV       & 10.59\% & 87.17\% &  9.61\% & 90.13\% &  9.23\% & 91.93\% &  8.14\% & 77.55\% \\
			\midrule
			& \multicolumn{2}{c}{\textbf{AUC}} 
			& \multicolumn{2}{c}{\textbf{AUC}} 
			& \multicolumn{2}{c}{\textbf{AUC}} 
			& \multicolumn{2}{c}{\textbf{AUC}} \\
			\midrule
			STRIP & \multicolumn{2}{c}{0.570} & \multicolumn{2}{c}{0.546} & \multicolumn{2}{c}{0.469} & \multicolumn{2}{c}{0.465} \\
			Teco  & \multicolumn{2}{c}{0.827} & \multicolumn{2}{c}{0.837} & \multicolumn{2}{c}{0.818} & \multicolumn{2}{c}{0.744} \\
			\bottomrule
		\end{tabular}
	}
	\vspace{1em}
\end{small}
\end{table}

\textbf{Evaluation on High-Resolution Dataset.} We conduct a group of experiments using the high-resolution dataset Tiny ImageNet, and list the results in Table \ref{tab:Tiny_Imagenet}, from which we observe our SifterNet remains at a considerable level in backdoor removal rate, compared to the low-resolution datasets like CIFAR-10 and GTSRB. However, there is a noticeable drop in the target model's accuracy on clean samples after purification process by SifterNet. We think the main reason lies in the increased number of image categories, which reduces the memory capacity and recall accuracy of the trained Hopfield network, which results in a larger discrepancy between our SifterNet-processed images and the original clean ones, thereby reducing target model's recognition accuracy. Nevertheless, while compared to other backdoor purification and detection methods, it is evident that these backdoor-removal methods would become more sensitive to high-resolution images. Even so, our proposed SifterNet still demonstrates superior performance compared to IFMV.

\textbf{Evaluation with Vision Large Model.} Still using Tiny ImageNet, we also evaluated the performance with the large-scale vision model ViT\_b\_16, and experimental results are demonstrated in Table \ref{tab:ViT}, from which we find that the effectiveness of our SifterNet shows a certain decline compared to that using dataset CIFAR-10 and the target model ResNet18. 
We think the behind reason lies in that ViT\_b\_16 model is more vulnerable than the convolution neural network ResNet18, stemming from the tight-correlation between segmented pixel patches in Transformer architecture in vision large model. After the image undergoes binarization and Hopfield-based recall, it can be deemed as introducing somewhat perturbation into this image, leading to a more significant drop in recognition accuracy. Of course, a target model's sensitivity to high-resolution image's perturbation is also an important factor to judge the performance of backdoor removal methods. At same time, we observe IFMV also drop much more and performs worse than SifterNet in both $Acc.$ and $ASR$ metrics. In contrast, the backdoor detection methods tend to be more stable in comparison, this is because they do not modify/eliminate the adversarial pixels.

To sum up, our proposed approach aims to purify trigger without considering whatever the target models are. Thereby, it is effective to both active and passive backdoor attacks. Furthermore, due to the constraint of "attractor domain", it can always converge to minimum energy over time, i.e. purify the trigger in a proper manner under the guidance of memorization and association procedure, giving rise to an adequate trigger purification to a large extent.

\begin{table}[tbhp]
	\centering
	\caption{Defense performance tested on Cifar10 ViT b 16 model}
	\label{tab:ViT}
	\begin{small}
		%	\vspace{1em}
		\begin{tabular}{lcccc}
			%	\begin{tabular}{p{1.1cm}>{\centering\arraybackslash}p{1.32cm}>{\centering\arraybackslash}p{1.32cm}>{\centering\arraybackslash}p{1.32cm}>{\centering\arraybackslash}p{1.32cm}>{\centering\arraybackslash}p{1.32cm}>{\centering\arraybackslash}p{1.32cm}}	
				\toprule
				\multirow{2}{*}{\textbf{Defense}} &
				\multicolumn{2}{c}{\textbf{BadNet}} &
				\multicolumn{2}{c}{\textbf{Blended}} \\
				\cmidrule(lr){2-3}\cmidrule(lr){4-5}
				& Acc. & ASR & Acc. & ASR \\
				\midrule
				Clean     & 95.93\% & 94.46\% & 96.90\% & 99.60\% \\
				SifterNet & 34.62\%$\pm$2\% & 25.91\%$\pm$4\% & 32.53\%$\pm$1\% & 29.33\%$\pm$5\% \\
				IFMV      & 11.70\% & 94.40\% & 15.44\% & 66.81\% \\
				\midrule
				& \multicolumn{2}{c}{\textbf{AUC}} & \multicolumn{2}{c}{\textbf{AUC}} \\
				\midrule  % 中间的横线
				STRIP     & \multicolumn{2}{c}{0.456} & \multicolumn{2}{c}{0.546} \\
				TeCo      & \multicolumn{2}{c}{0.842} & \multicolumn{2}{c}{0.436} \\
				\bottomrule
			\end{tabular}
			
			%	\vspace{1em}
			
			%—— 第二部分（后半截列） ——
			\begin{tabular}{lcccc}
				%	\begin{tabular}{p{1.1cm}>{\centering\arraybackslash}p{1.32cm}>{\centering\arraybackslash}p{1.32cm}>{\centering\arraybackslash}p{1.32cm}>{\centering\arraybackslash}p{1.32cm}}	
					\toprule
					\multirow{2}{*}{\textbf{Defense}} &
					\multicolumn{2}{c}{\textbf{SIG}} &
					\multicolumn{2}{c}{\textbf{WaNet}} \\
					\cmidrule(lr){2-3}\cmidrule(lr){4-5}
					& Acc. & ASR & Acc. & ASR \\
					\midrule
					Clean     & 86.50\% & 92.59\% & 89.20\% & 80.99\% \\
					SifterNet & 28.51\%$\pm$1\% & 27.46\%$\pm$1\% & 26.89\%$\pm$2\% &  9.63\%$\pm$5\% \\
					IFMV      & 15.80\% & 92.20\% &  15.55\% & 0.66\% \\
					\midrule
					& \multicolumn{2}{c}{\textbf{AUC}} & \multicolumn{2}{c}{\textbf{AUC}} \\
					\midrule  % 中间的横线
					STRIP     & \multicolumn{2}{c}{0.114} & \multicolumn{2}{c}{0.431} \\
					TeCo      & \multicolumn{2}{c}{0.746} & \multicolumn{2}{c}{0.648} \\
					\bottomrule
				\end{tabular}
			\end{small}
			%	\vspace{1em}
		\end{table}
	
\subsection{Computation complexity} \label{Sec:timeOverhead}
Hopfield network is in essence a single-layer fully connected neural network, thus theoretically, it ought to possess 1024 neurons towards processing CIFAR-10 and GTSRB datasets, the corresponding FLOPs of the network is 2,097,152. %, and the time complexity is $O(n*n)$.
If it has 784 neurons corresponding to MNIST and FashionMNIST datasets, the corresponding FLOPs is 1,229,312. The time complexity is $O(n*n)$. %as well

\begin{figure} [tbhp]
\centering
\subfigure[Serial inference]{
%	\subfloat[2D Ising model with lattice arrangement]{
	\begin{minipage}[t]{0.48\columnwidth}
		\centering
		\includegraphics[width=2.6in, height=1.4in]{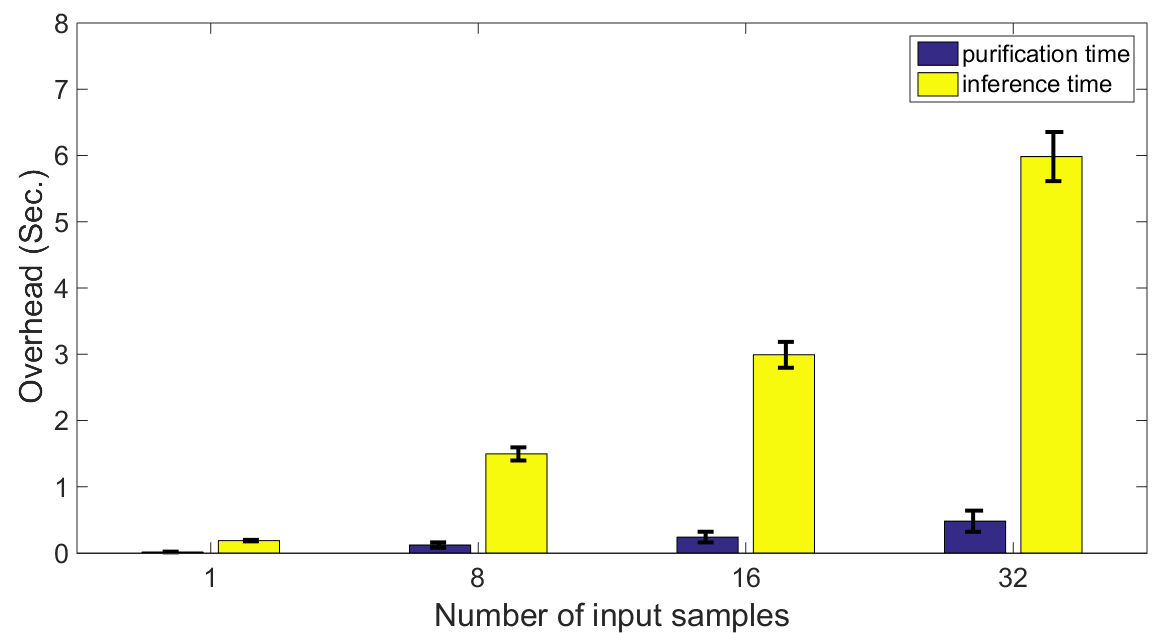}
		%	\caption{xx}
		\label{fig:serialTimeOverhead}
	\end{minipage}
}
\vspace{0.02cm}
%     \quad
\subfigure[Parallel inference]{
	%	\subfloat[Black-white image represented by magnetic pins]{
		\begin{minipage}[t]{0.48\columnwidth}
			\centering
			\includegraphics[width=2.6in, height=1.4in]{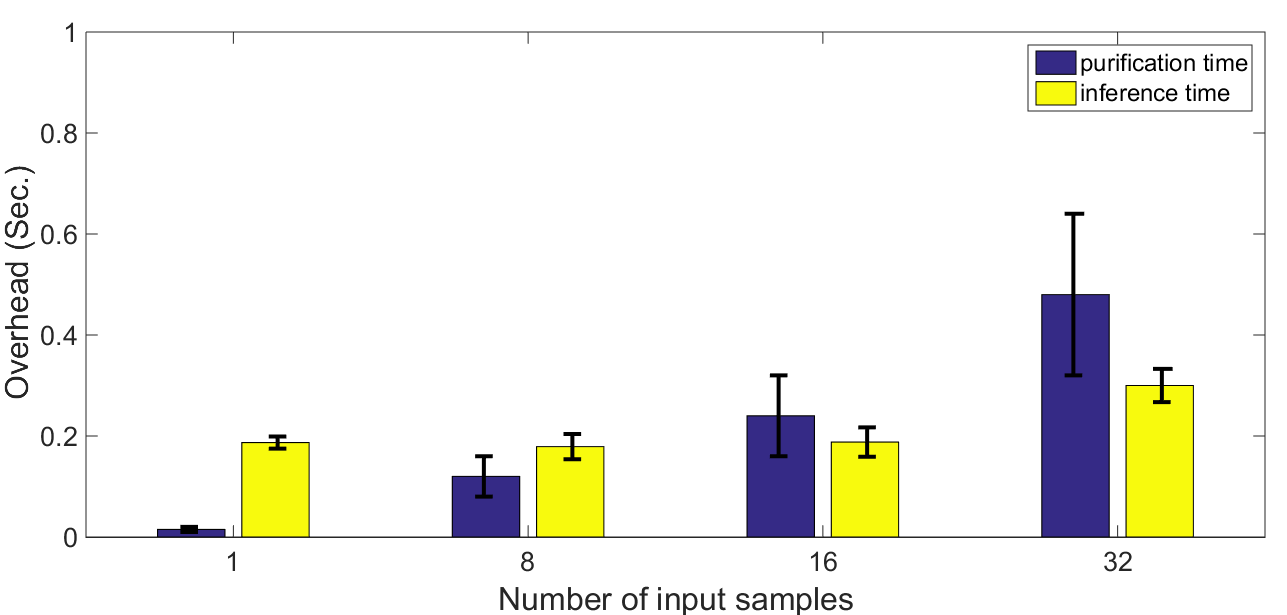}
			%	\caption{xx}
			\label{fig:parallelTimeOverhead} 
		\end{minipage}
	}
	\caption{The time overhead under serial and parallel computations.}
	\label{fig:timeOverhead}
\end{figure}

Lightweight is another advantage of our approach, resulting from the simple structure of Hopfield network with only one single-layer fully-connected neural network. To verify this, we mainly divide the computation overhead into two parts: trigger-purification time and input inference time. On dataset GTSRB, a group of experiments are run in five times using our self-built neural network models, and then calculate the standard error of the mean. The results are drawn in Fig. \ref{fig:timeOverhead} in terms of purification time and inference time under serial and parallel computations, from Fig. \ref{fig:timeOverhead}(a) we can observe that: i) the overhead of trigger purification only takes a small part compared to the inference part, and the tendency becomes more sharply, i.e the overhead of inference part costs more and more as the number of input samples increases, this indicates our Hopfield network-oriented purification is efficient; ii) to alleviate the time overhead of inference phase, we also introduce the parallel computation, i.e. simultaneously input multiple samples into Hopfield network for purification, from Fig. \ref{fig:timeOverhead}(b) we can see the inference time dramatically decreases, even less than purification phase as the number of input samples enlarges. In a nutshell, our SifterNet achieves lightweight computation.   

Moreover, we also conduct the time overhead test on the open-source BackdoorBench. The testing environment is as follows: CPU: 14 vCPUs Intel(R) Xeon(R) Platinum 8362 @ 2.80GHz, GPU: RTX 3090 (24GB) x1, RAM: 45GB. Regarding the detection time for all trigger-implanted samples on GTSRB dataset, STRIP takes 67.62 seconds, TeCo takes 2142.23 seconds, and our SifterNet takes 56.84 seconds. Both our method and STRIP are relatively lightweight, providing trigger-purification and trigger-detection results quickly, leading to the defense time being approximately 3 times the inference time of the target model. On the other hand, TeCo, due to its non-targeted detection that relies on multiple and varied disturbances to check whether the image contains a trigger, requires more computational cost. As a result, the defense time of TeCo is significantly longer than that of our method and STRIP, approximately 95 times the inference time of target model.

We also measure the detection time for a defense method similar to the filtering-based defense approach, IFMV. TeCo took 215.6 seconds to process the unknown data in the test set of GTSRB dataset, which is approximately 11 times the inference time of the target model. Since IFMV also requires generating multiple sets of disturbance samples to perform voting for the final result, its time and computational overhead are relatively significant.

%then calculate the ratio of the two parts. For GTSRB dataset, our SifterNet performs 0.002 seconds to process the trigger for an input while 0.137 second to execute inference. Furthermore, from the viewpoint of comparison, the quantitative standard is utilized, and our SifterNet has a time ratio close to 1, compared to the ratios of 1.5\textasciitilde3 in SentiNet, 1.75\textasciitilde6 in STRIP and above 23 in IFMV. 

\section{Robustness enhancement}\label{Sec:robustEnhancement}	
\subsubsection{Hopfield-network expansion}
Memorization and association strictly depend on the capacity of Hopfield network. The more the network capacity, the more the images can be processed in a batch. Theoretically, the capacity of Hopfield network with $n$ neurons is \(C \cong {\raise0.7ex\hbox{$n$} \!\mathord{\left/{\vphantom {n {2\log _2 n}}}\right.\kern-\nulldelimiterspace} \!\lower0.7ex\hbox{${2\log _2 n}$}}\) \cite{Liou1999, Liou2006}. However, pursuant to our experiments with MNIST dataset, its capacity can only reach up to 10\%$\sim$15\% of the theoretical value. when the number of memorized images is beyond this range, the association effect starts to deviate. 

\begin{figure}[tbhp]
	\centering
	\includegraphics[width=\columnwidth]{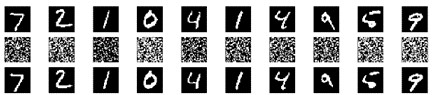}
	\caption{Sample break up and recovery (up: original, middle: scattered, bottom: recovered).}
	\label{fig:capacityExtension}
\end{figure}

\begin{figure}[tbhp]
	\centering
	\includegraphics[width=3.0in, height=1.4in]{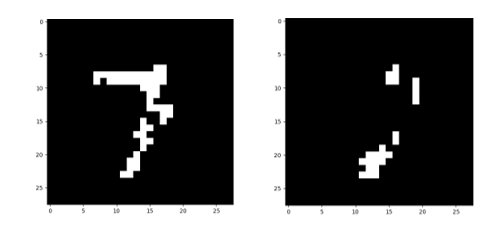}
	\caption{Recovered "7" (left: with capacity extension, right: without).}
	\label{fig:handwrittenRecover}
\end{figure}

Moreover, the more the images are memorized, the worse the association effect becomes. This is because the main-semantics region concentrates on the middle of image, which renders the attractors are too close to remember enough images. To surround this problem, we figure out a neural network-assistant solution to promote Hopfield network's capacity, i.e. mess image and scatter semantics across diverse parts over the entire region. For detail, as shown in Fig. \ref{fig:capacityExtension}, we employ neuronal network to create a set of mapping functions that enables to break up and recovers the samples under the assistance of capacity extension. At memorization phase, the input sample is mapped and scrambled at first, then Hopfield network is leveraged to remember the scrambled image. At association phase, the input is also scrambled at the beginning, and then the scrambled image is sent into Hopfield network for purification. Finally, an "recall" image is recovered by the mapping relationship. Fig. \ref{fig:handwrittenRecover} exhibits two handwritten digits "7" generated through the association of Hopfield network, the left is with the capacity-extension operation, while the right is without. Obviously, the left is closer to the original image, and looks more natural than the right.

\subsubsection{Data augmentation}
\begin{figure}[ht]
	\centering
	\includegraphics[width=\columnwidth]{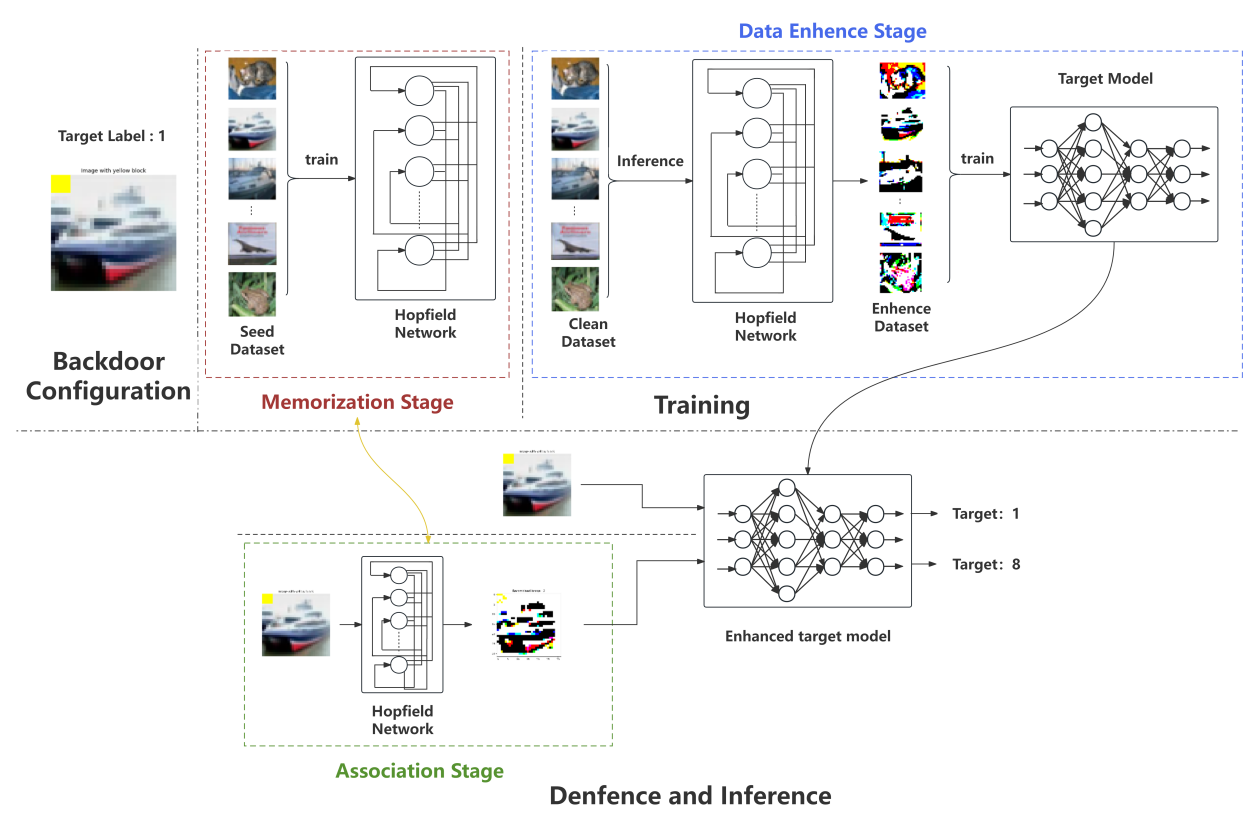}
	\caption{Framework of data augmentation.}
	\label{fig:dataAugmentation}
\end{figure}

This part discusses the accuracy enhancement from the perspective of training data, i.e. data augmentation. The core idea is to adopt the samples that processed by Hopfield network in advance to retrain model, with the purpose of declining the probability of misclassification of the deformed/purified samples. Fig. \ref{fig:dataAugmentation} sketches the overall framework of data augmentation. In addition, Table \ref{table:dataAugmentation} also demonstrates the experimental results with/without data augmentation. The experimental results are listed in Table \ref{table:dataAugmentation} with datasets CIFAR-10 and GTSRB, from which we can observe the performance has an obvious improvement both in $Acc.$ and $ASR$, for example, our SifterNet enables to decline the $ASR$ from 39.02\% to 32.38\% on CIFAR-10, and from 34.38\% to 29.00\% on GTSRB, simultaneously, the routine accuracy increases by 3.89\% on CIFAR-10, and 3.08\% on GTSRB. Therefore, the accuracy promotion unveils such data augmentation fashion can indeed mitigate the trouble of sample information loss caused by the affection of purification to some extent. 

\begin{table}[t]
	\centering
	\caption{The performance before and after data augmentation}
	\label{table:dataAugmentation}
	\begin{small}
		%\vspace{1em}
		\begin{tabular}{lcccccc}  % 使用标准的列格式
			\toprule  % 表格上方的线
			\textbf{SifterNet} & \multicolumn{2}{c}{\textbf{Before Data Augmentation}} & \multicolumn{2}{c}{\textbf{After Data Augmentation}} \\
			\midrule  % 中间的横线
			& \textbf{Acc.} & \textbf{ASR} & \textbf{Acc.} & \textbf{ASR} \\
			\midrule  % 中间的横线
			CIFAR-10 & 36.06\%$\pm$6.50\% & 39.02\%$\pm$6.00\% & 39.95\%$\pm$4.00\% (\textcolor{red}{3.89\%$\uparrow$}) & 32.38\%$\pm$5.00\% (\textcolor{red}{6.64\%$\downarrow$}) \\
			GTSRB & 91.75\%$\pm$5.00\% & 34.38\%$\pm$5.00\% & 94.83\%$\pm$3.50\% (\textcolor{red}{3.08\%$\uparrow$}) & 29.00\%$\pm$6.50\% (\textcolor{red}{5.38\%$\downarrow$}) \\
			\bottomrule  % 表格下方的线
		\end{tabular}
		%\vspace{1em}
	\end{small}
\end{table}

\section{Related work} \label{Sec: RelatedWork}
To date, there are several sublines of backdoor defense in terms of trigger treatment. The representative approach STRIP \cite{STRIP} uses the thought of stacking image to check backdoor in a black-box means, however, its drawback lies in the incapability of handling active backdoor attacks, stemming from the dependence on observation of target model’s outputs. Thereby, attackers can easily bypass its detection through tweaking manipulation, i.e. attackers enable to prevent defender from finding out common adversarial features. Differently, our proposed SifterNet is data-oriented and does not require the knowledge of model’s output or features, it has more generalization and adaptability disregarding whatever the passive or active attacks. 

Similar to ours, SentiNet \cite{SentiNet} is a data-oriented countermeasure that aims at identifying the potential triggers. Nevertheless, it assumes the implanted triggers possess obviously recognizable pattern. This assumption significantly constrains its application, due to the fact that the current sophisticated backdoor attacks can generate scattered triggers without visual perturbation in a covert and subtle means. IFMV \cite{IFMV} is also a data-oriented black-box defense method through adding noise to purify triggers in an inverse thought. However, this noise-addition orientation is usually deviated and hard to exactly locate the trigger area, which easily causes the strict damage to both trigger and core semantics. It tends to be capable to tackle the simple-semantics data, but struggle with complex-semantics data, this well explains why the performance of three-channel CIFAR-10 has noticeable declined compared to that of single-channel MNIST. As for the trigger-orientation problem, our SifterNet can properly solve it through resorting to the memorization-association functionality of Hopfield network.  

On the other hand, the model-oriented defense studies \cite{Neural_Cleanse, BayBFed, Demon_in_the_Variant, FLAME} are also proposed, aiming at retraining neural networks to discard backdoor pattern. For detail, targeted datasets can be specially created to fine-tune target models with the purpose of network purification. Thus, this line of work is usually white-box and requires the beforehand knowledge of attack activities, which strictly constrains the practical application. Easy to know, this line of work would be time-consuming due to the requirement of model retraining. In reality, computational power is a critical factor during the deployment, thereby, an appropriate countermeasure might be a lightweight data-oriented trigger purification mechanism.

\section{Discussion} \label{Sec: OpenIssue}
In the process of executing our experiments, we also find several issues: i) \textbf{Sensitive to Vision Large Model.} Although our SifterNet can significantly reduce the attack success rate of backdoor, it is sensitive to the Transformer architecture-based vision large model, e.g. ViT, we think the behind reason may lie in that the purification process brings negative affection on the tight-correlation of segmented pixel-patched by the Transformer, which furtherly influences the correct semantics learning, finally leading to the reduction of accuracy; and ii) \textbf{Large Number of Categories of Dataset.} As we know, the routine Hopfield network is in general small with only one-layer , thus it has limited capacity, a large number of categories can degrade the memory-recall functionality. That is to say, a fixed Hopfield network cannot remember or recall input data well once the number of categories exceed its capacity. To solve this problem, we need to expand the capacity of Hopfield network. In our work, we resort to an assistant neural network to expend the capacity.

\end{document}